\documentclass[10pt,twocolumn,letterpaper]{article}
\usepackage{caption}
\usepackage{packages/iccv}
\usepackage{times}
\usepackage{epsfig}
\usepackage{graphbox} %
\usepackage[export]{adjustbox}
\usepackage{amsmath}
\usepackage{amssymb}
\usepackage{float}
\usepackage{color, colortbl}
\usepackage{booktabs}
\usepackage{authblk}
\usepackage[pagebackref=true,breaklinks=true,letterpaper=true,colorlinks,bookmarks=false]{hyperref}

\usepackage[utf8]{inputenc} 
\usepackage[T1]{fontenc}    
\usepackage{hyperref}       
\usepackage{url}            
\usepackage{booktabs}       

\usepackage{amsfonts}       
\usepackage{amssymb}
\usepackage{amsmath}
\usepackage{amsthm}
\usepackage{bbm}
\usepackage{ulem}

\usepackage{nicefrac}       
\usepackage{microtype}      
\usepackage{lipsum}		
\usepackage{graphicx}
\usepackage{doi}

\usepackage{xcolor}
\usepackage{mdframed}
\usepackage{hhline}

\usepackage{graphicx}

\usepackage[font=small,labelfont=bf]{caption}

\definecolor{c1}{gray}{0.96}
\definecolor{c2}{gray}{0.70}
\definecolor{c3}{gray}{0.40}
\definecolor{c4}{gray}{0.48}
\definecolor{hlcolor}{rgb}{0.55,0.2,0.3}
\definecolor{emphcolor}{rgb}{0.4,0.6,0.6}

\hypersetup{
    colorlinks=true,       
    linkcolor=red,          
    citecolor=green,        
    filecolor=magenta,      
    urlcolor=emphcolor      
}

\newcommand{\txtbold}[1]{\textbf{{\sffamily
\color{c3}{#1}}}}

\newcommand{\figcaption}[3]{
\begin{figure}[ht]
  \centering
  \includegraphics[width=\linewidth]{#1}
  \caption{#2}
  \label{#3}
\end{figure}
}

\theoremstyle{definition}

\mdfdefinestyle{mdlight}{
  hidealllines=true,backgroundcolor=c1,
  innerleftmargin=8pt,innerrightmargin=8pt,innertopmargin=8pt,
  rightmargin=10pt,skipabove=0,skipbelow=0}
\mdfdefinestyle{mddark}{
  hidealllines=true,backgroundcolor=c2,
  innerleftmargin=15pt,innerrightmargin=15pt,innertopmargin=8pt,
  rightmargin=5pt,}
\surroundwithmdframed[style=mdlight]{ppr}
\surroundwithmdframed[style=mdlight]{dsc}

\newcommand{\pref}[3]{\begin{mdframed}[style=mdlight]
\txtbold{\href{#2}{[x]} #1}}

\newcommand{\eqnsplit}[1]{\begin{equation*}\begin{split}#1\end{split}\end{equation*}}

\newcommand{\nrm}[1]{\vert\vert #1\vert\vert}

\newcommand{\set}[1]{\{#1\}}

\newcommand{\R}{\mathbb{R}}

\iccvfinalcopy
\begin{document}
\title{\vspace{-1.5em}Leveraging Deepfakes to Close the Domain Gap between \\Real and Synthetic Images in Facial Capture Pipelines}
\author[1]{Winnie Lin}
\author[1,2]{Yilin Zhu}
\author[1,2]{Demi Guo}
\author[1,2]{Ron Fedkiw}
\affil[1]{Stanford University}
\affil[2]{Epic Games}

\twocolumn[{%
\renewcommand\twocolumn[1][]{#1}%
\maketitle
\begin{center}
    \centering
    \vspace{-1.5em}  \includegraphics[width=\textwidth]{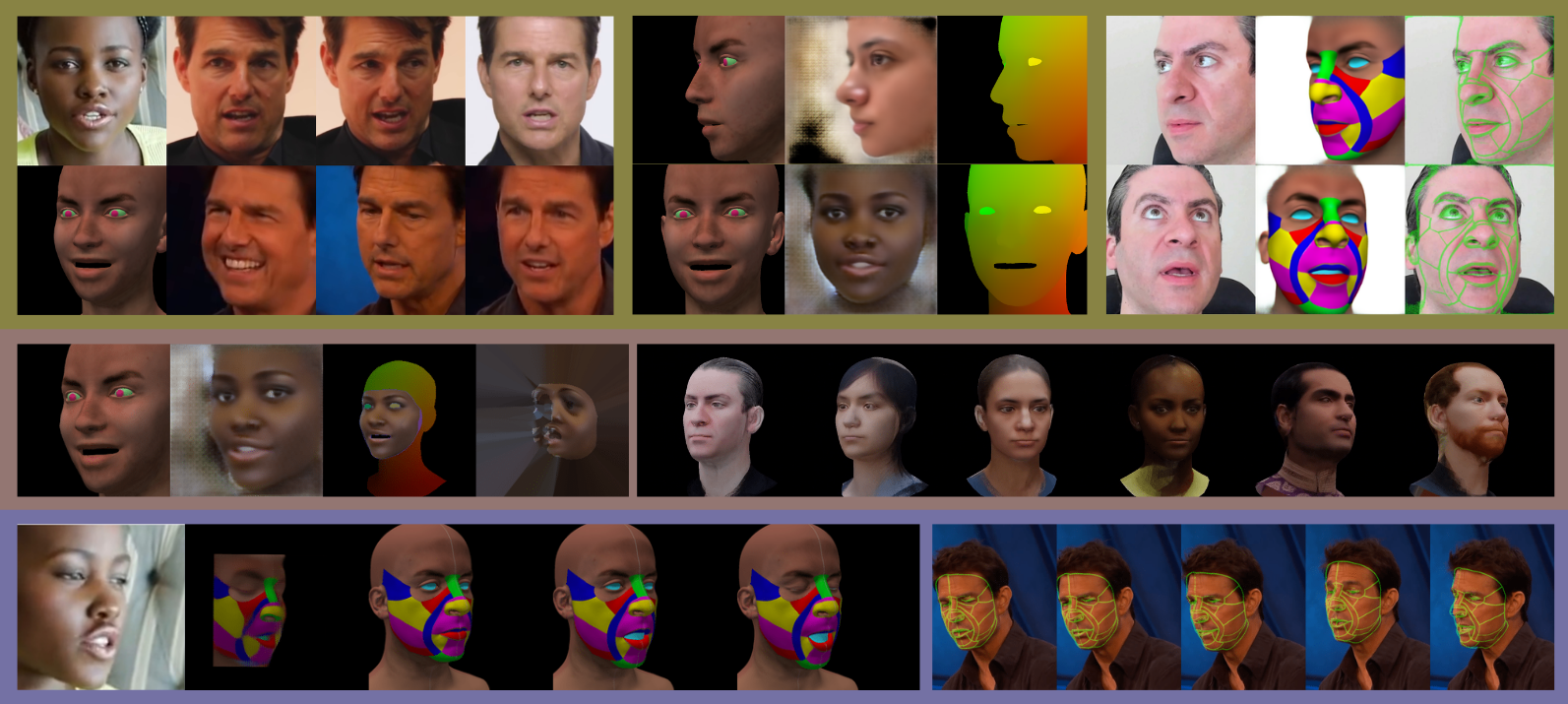}
  \captionof{figure}{Top left: Given either real or synthetic input images, our automatic data curation allows us to find similar pose and expressions from an in-the-wild image dataset. This allows us to train personalized networks in a scalable way, requiring only a few hundred to a few thousand in-the-wild images collected via cellphones, webcams, or youtube videos. Top middle: The ability to inference from synthetic to real is used in our appearance capture pipeline. Top right: The ability to inference from real to synthetic is used in our motion tracking pipeline.
  Middle row: Our appearance capture pipeline, along with some representative results (on various races). Third row: Our motion capture pipeline, illustrating results from both inverse rendering and regression.}
\end{center}}]
\begin{abstract}
We propose an end-to-end pipeline for both building and tracking 3D facial models from personalized in-the-wild (cellphone, webcam, youtube clips, etc.) video data.
First, we present a method for automatic data curation and retrieval based on a hierarchical clustering framework typical of collision detection algorithms in traditional computer graphics pipelines. Subsequently, we utilize synthetic turntables and leverage deepfake technology in order to build a synthetic multi-view stereo pipeline for appearance capture that is robust to imperfect synthetic geometry and image misalignment. The resulting model is fit with an animation rig, which is then used to track facial performances. Notably, our novel use of deepfake technology enables us to perform robust tracking of in-the-wild data using differentiable renderers despite a significant synthetic-to-real domain gap. Finally, we outline how we train a motion capture regressor, leveraging the aforementioned techniques to avoid the need for real-world ground truth data and/or a high-end calibrated camera capture setup.
\end{abstract}

\section{Introduction}
There is a growing interest in avatar personalization and facial motion tracking for consumer use, as avatars that accurately capture the likeness, appearance, and motion of the user increase user engagement and immersiveness. Accordingly, there is a plethora of industry interest in 3D avatar customization as part of a greater push for immersive entertainment and community building, as demonstrated by the popularity of Roblox, Unreal Engine's Metahuman creator, and various virtual/augmented reality related avatar personalization endeavours from Meta, Apple, and others. Even when a personalized avatar is not desired, they are quite useful as strong priors for the motion capture required for puppeteering.
\par Highly-specialized high-end methods such as lightstage capture \cite{debevec_acquiring_2000}, multi-view stereo reconstruction \cite{beeler2011high}, head-mounted cameras \cite{bhat2013high} etc. have an important role in the special effects industry, enabling the acquisition of appearance, geometry, and motion from image (and marker) data, but these pipelines are computationally and resource intensive, and require carefully calibrated hardware, domain expertise, and manual intervention to operate. 
Meanwhile, with the advent of data-driven deep-learning methods, there is an exciting surge in the democratization of avatar creation and generation. Although this is still an actively developing area of research with many open-ended questions, it is undeniable that building semi-automated pipelines that do not require specialized setup and expertise is a key step to creating personalized avatars at scale.
\par Fully data-driven deep-learning methods often suffer from either single-view based input constraints, data diversity problems, or both. Single-view based appearance and motion capture is a wildly underconstrained problem, and it is still unclear how one would best incorporate arbitrary multi-view in-the-wild imagery into a pure deep-learning approach. The overreliance on large datasets gives rise to many practical problems regarding data quality, diversity, and bias. Although this is a common issue for many deep-learning applications, data diversity is particularly crucial when personalization/customization is of key importance. For example, papers in the AI ethics community \cite{buolamwini2018gender} have pointed out that facial models trained on Caucasian-centric (or even worse, Caucasian male-centric) data perform poorly and unreliably on other ethnic groups. Problems like this are inherent in most state-of-the-art methods, and need to be adequately addressed \cite{khan2021one}\cite{mitchell2019model} before such methods are able to be deployed at scale.
To mitigate these issues, our approach overfits small, specialized deep-learning models to each specific subject of interest. In order to make such a method tractable and scalable enough for democratization, we automate the curation and compactification of personalized input data, relying on our proposed data curation algorithms to aggressively prune the data and thus reduce the compute time needed to train specialized deep neural networks.
\par Our pipeline aims to capture the strengths of both traditional methods and deep-learning methods. More specifically, for appearance capture, we rely on classical multi-view stereo techniques to be less beholden to the monocular-view constraints of deep-learning approaches, while utilizing synthetic turntables and leveraging deepfake technology to mitigate multi-view stereo misalignment and obviate the need for any specialized hardware. For motion capture, we utilize deepfake technology to close the domain gap between real in-the-wild images and synthetically-generated data, facilitating robust inverse-rendering even in the absence of photorealistic appearance and lighting models. 
In addition, a similar leveraging of deepfake technology allows us to train a fast and efficient motion capture regressor in the absence of ground truth control parameters for real imagery.
\section{Prior Work}
Existing approaches for appearance and motion capture typically fall into two categories: either high-end personalized approaches widely used in the visual effects industry or democratized monolithic approaches widely used in commercial settings. Here, we briefly summarize some relevant work, referring the interested reader to \cite{zollhofer2018state} for a more in-depth survey.
\par \textbf{High-end methods:} In \cite{beeler2010high}, the authors built a multi-view stereo pipeline carefully optimized for geometry capture, reconstructing pore-level detail through iterative stereo-pair based geometric reconstruction. In \cite{debevec_acquiring_2000}, the authors built a controlled spherical lighting setup for high quality capture of the reflectance model of a face. Both of these methods and their successors (e.g. \cite{ghosh2011multiview}\cite{riviere2020}) are an integral component of appearance capture pipelines in most major visual effects studios \cite{debevec2012light}\cite{hendler2018avengers}\cite{medusa}. 
\par Early works on high-end motion capture reconstructed dense mesh sequences from video, with methods such as \cite{zhang-siggraph2004-stfaces} and \cite{debevecmocap} employing multi-view stereo and scanline projectors. Current high-end motion capture approaches typically require both head-mounted cameras and physical markers placed on the face (see e.g. \cite{bhat2013}\cite{moser2017masquerade}\cite{mocapspotlight}). Solving the capture problem with sparse markers necessitates either a personalized animation rig (created via multi-view stereo \cite{beeler2011high}, deformation transfer \cite{sumner2004deformation}, artist supervision, etc.) or a data-driven approach (e.g. \cite{cong2019} interpolates from reconstructed and/or simulated geometry, \cite{hendler2018avengers} performs dense marker driven mesh deformations, etc.). Markerless methods, previously more commonly utilized in democratized approaches, have in recent years also become an active area of research for high-end applications, with inverse rendering technology (e.g. \cite{ravi2020pytorch3d}\cite{opendr}) being used to solve for animation parameters directly from image data (e.g. \cite{bao2019}\cite{flux}).
\par \textbf{Democratized methods:} These methods typically start with a parametrized template geometry/texture usable across all subjects. \cite{blanz1999morphable} pioneered a statistical PCA approach, generating a linear parametrization from a dataset of scans. This approach has been the standard for many years, with gradually increasing dataset size and complexity \cite{booth20163d}\cite{li2017learning}. More recently, deep-learning approaches (which generate nonlinear parametrizations) have gained popularity (e.g. \cite{vae3dmesh}\cite{meshgan}), and hybrid PCA/deep-learning methods \cite{yang2020facescape} have also emerged. Both the PCA and deep-learning approaches typically parameterize both appearance and expression \cite{egger20203d}; as such, they can be utilized for both appearance and motion capture. 
\par To perform appearance capture, it is common to utilize facial landmarks (e.g. \cite{bulat2017far}) or other sparse features as loss contraints for training regressors that predict geometry parameters from monocular images (e.g. \cite{sanyal2019learning}\cite{tewari2018high}). Democratized motion capture approaches typically utilize low-dimensional parametrizations of geometry, either data-driven (e.g. \cite{blanz1999morphable}\cite{vae3dmesh}) or artist-sculpted (e.g. \cite{blendshapes}). The parameters are typically determined using optimization to fit the geometry directly to landmarks \cite{chrysos2018comprehensive} or depth maps \cite{weise2011realtime}\cite{li2013realtime}. Alternatively, monolithic deep-learning regressors can be trained via sparse landmark constraints to directly output parameters from images \cite{liu2017dense}\cite{Tewari_2018_CVPR}\cite{navarro2021fast}. A recent example of a regressor based approach \cite{sanyal2019learning} jointly predicts appearance and pose/expression in their end-to-end model, utilizing a landmark loss and a mesh based triplet loss during training.
\par\textbf{Hybrid approaches:} The development of commercially available AR/VR headsets has led to research on democratized markerless motion capture for head-mounted cameras, see e.g. \cite{pixelcodec}\cite{facevr}. These methods typically use monolithic multi-identity neural networks, although \cite{jourabloo2021} adds a personalized conditioner by inputting a neutral face mesh of the subject. Our approaches to both appearance and motion capture similarly combine high-end and democratized approaches. We aim to build personalized pipelines typical of high-end applications while also dealing with the uncertainties of in-the-wild cameras and data. 
\par Notably, \cite{moser2021semi} (contemporaneous with our work) takes a similar approach to training personalized motion capture regressors, learning a joint embedding space between real and synthetic images for markerless tracking in a ``green screen" capture setup. Although their approach does not handle in-the-wild data, it does show promising generalizability to less-constrained data capture. In addition, our work on the motion capture regressor is heavily related to and preceded by \cite{deepappearancemodels}, which utilizes joint embeddings between real and synthetic images to do markerless motion capture with a multi-view head-mounted camera; however, their controlled capture setup is more stringent than what is required of our approach (and \cite{moser2021semi}). Our use of inverse rendering for offline motion capture also bears similarity to \cite{gecer2019ganfit} and \cite{Tewari_2018_CVPR}. \cite{gecer2019ganfit} describes a single-image based appearance capture pipeline, and utilizes a neural renderer for offline optimization.  \cite{Tewari_2018_CVPR} uses a neural renderer to learn unconstrained deformable motion from in-the-wild images. In comparison, our inverse-rendering approach differentiates through a dense pixel loss instead of sparse keypoints and is a single-identity model trained only on subject-specific data, but unlike \cite{Tewari_2018_CVPR} our method is currently not designed to run in real-time.
\par Our approach to appearance capture is similar in spirit to \cite{ichim2015dynamic}\cite{Luo_2021_CVPR} \cite{deepfacenormalization}. \cite{ichim2015dynamic} utilizes a traditional model fitting approach that generates avatars from a short video clip of a single subject; in our approach, we obviate the need for camera estimation and image alignment by utilizing deepfake networks. \cite{Luo_2021_CVPR} achieves high-quality results on geometry/texture acquisition via a multi-identity GAN, similarly using dramatically less data than other contemporary works through careful selection and design of their dataset. \cite{deepfacenormalization} builds a multi-identity pipeline for face frontalization, where an image of a frontal face with diffuse lighting is generated from inputs consisting of arbitrary pose/expression and lighting. While leveraging similar concepts to \cite{Luo_2021_CVPR}\cite{deepfacenormalization}, our method focuses on using a collection of images of a single subject instead of single images of multiple subjects.
\section{Hierarchical Clustering for Data curation}
A crucial part of both our appearance capture and motion capture pipelines is the curation of data from in-the-wild footage. The importance of data curation in our approach is twofold: We aim to not only prune irrelevant and noise-inducing data, but also to craft compact datasets with the exact amount of information required for a given application. As shown in section 5, our data curation allows us to successfully train deepfake networks from scratch with fewer than 200 images in each dataset. In order to curate compact datasets, we devised a k-d tree based \cite{klosowski1998efficient} recursive partitioning of semantically meaningful image clusters (which additionally improves the efficiency of image retrieval.) 
Through iterative experimentation, we designed several features corresponding to pose, expression, and lighting:
\\\\\textit{Pose.} The 3-dimensional pose of a rigid object such as the skull is most commonly expressed in terms of a three degrees of freedom for translation and three degrees of freedom for rotation. However, when considering the projection of a rigid object into a 2D image, one can make some rough approximations in order to remove some degrees of freedom. 
Assuming minimal lens distortion (depth distortion, field of view dependencies, etc.) allows one to remove all three degrees of freedom corresponding to translation via image cropping and scaling. Similar assumptions allow one to remove rotation along the camera look-at axis, instead orienting the face to be upright via image rotation. This leaves two pronounced degrees of freedom, pitch and yaw, which both create significant occlusion of facial features.
\par Given an image $I$ of a face, we extract the pose feature $f_{pose}(I)$ as follows: we detect 68 2D facial landmarks $L(I){\in\R^{68\times2}}$ \cite{bulat2017far} and identify a subset of these landmarks $L_r {\in\R^{13\times2}}$ that don't vary too much with expression. We rigidly align $L_r$ to a predefined set of corresponding template landmarks $\tilde{L}_{r}$ that were derived assuming a frontal view with zero pitch and yaw. Even though $L_r$ mostly comes from images with nonzero pitch and yaw, this rigid alignment helps reduce dependencies on both scale and in-plane affine transformations enough to fit a 3D model to the result. That is, we can then determine the pitch and yaw parameters of a 3D template model that minimize the distance between the rigidly aligned $L_r$ and the 2D projection of corresponding markers embedded on the 3D model. Our pose feature $f_{pose}(I)$ is then the 2-dimensional vector corresponding to the resulting pitch and yaw of the 3-dimensional model. See Figure \ref{fig:pose}.
The least squares problem for finding the optimal rigid transformations for both steps can be solved using a simple singular value decomposition on a 3x3 matrix \cite{umeyama1991least}\cite{sorkine2009least}.
\figcaption{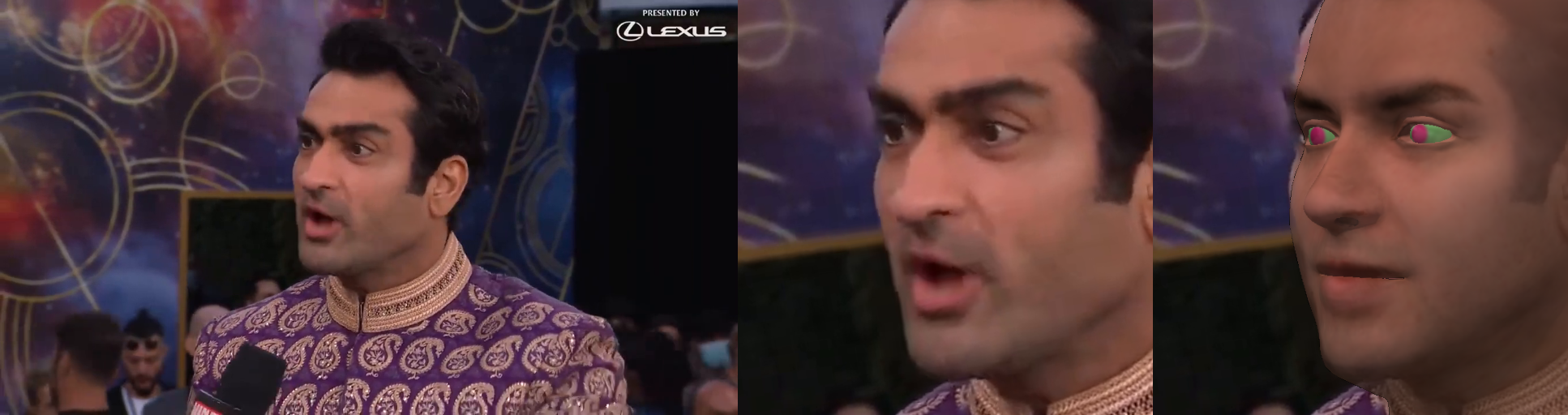}{From left to right: the image $I$, the image rigidly aligned to the set of template landmarks, a template 3D model rendered with the resulting pitch and yaw.}{fig:pose}
\\\\\textit{Expression.} We take landmarks $L_e\subset L(I)$ that correspond more heavily to expression (rather than pose), split them into (potentially overlapping) groups, and align each group separately to a 2D template $\tilde{L}_e$ in order to normalize for scale, in-plane rotation, and translation in a parts based manner. (Parts based models are common in human pose identification \cite{cheng2016person}, and bag-of-words approaches have also been utilized for facial recognition \cite{li2009facial}.) The results are then directly used as our feature vectors. See Figure \ref{fig:expression}. Of course, this high-dimensional feature vector could benefit from k-d tree based acceleration structures, but in practice our expression clusters are small enough that we leave their storage flat.
\figcaption{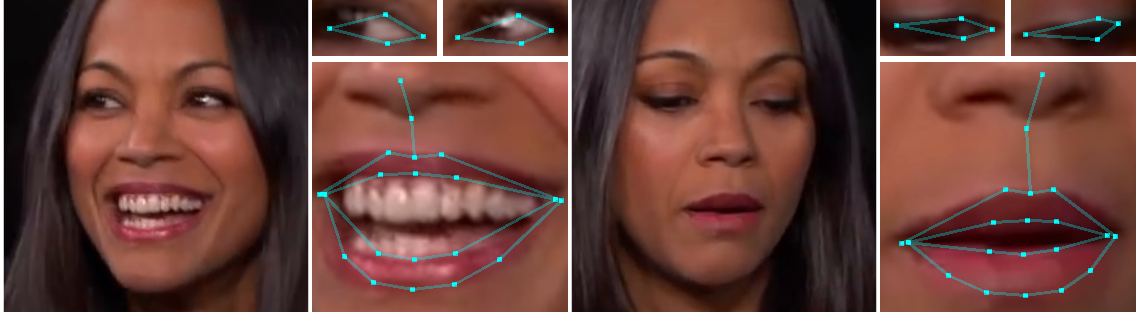}{
 Image $I$ and the aligned markers $f_{expression}(I)$ used in the expression matching layer (two examples are shown). To account for differences in global facial structure between a query subject and a dataset subject, we use a part based approach aligning the individual eyes and mouth separately.}{fig:expression}
\par An important technical detail is that the efficacy of $f_{expression}(I)$ hinges on an initial match based on pose, i.e. poor matches are common when the rigid poses are dissimilar. We address this by ensuring that $f_{expression}(I)$ only comes after $f_{pose}(I)$ in our hierarchical approach.
\\\\\textit{Lighting.} As opposed to the pose and expression vectors, which are mainly used to give accurate matches to an input image, our lighting based feature matching is meant to ensure variety and diversity in lighting conditions in order to counteract bias in our datasets.
Since the tip of the nose is highly rigid and contains a wide range of surface normal vectors, it well approximates environmental lighting conditions. Thus, motivated by the use of chrome spheres \cite{landis2002production}, we use a tight crop around the nose as a rough representation of the environmental lighting. The crop is rasterized to a $4\times3$ grid of RGB values, creating a feature vector $f_{lighting}(I)\in\R^{12\times 3}$. See Figure \ref{fig:light}.
\figcaption{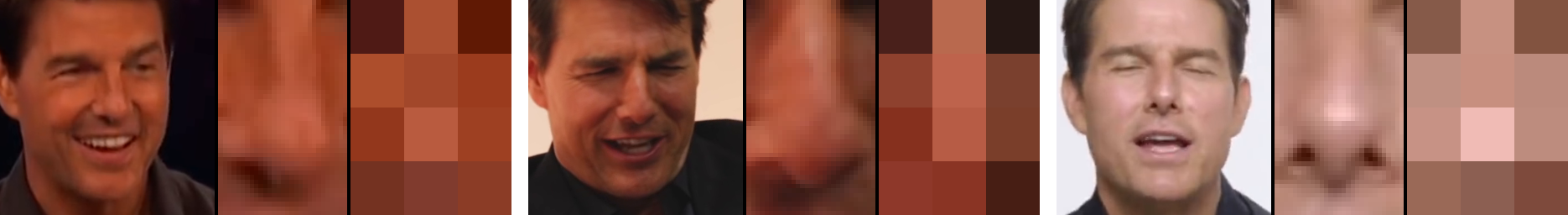}{Image $I$, tight nose crop of $I$, and rasterized crop that gets flattened to $f_{lighting}(I)$ (three examples are shown). Note that shadows (middle example) and specular highlights (rightmost example) are picked up in the rasterized features.}{fig:light}
\subsection{k-d Tree Construction} Given an image collection $\Omega=\set{I}$, one can evaluate a feature function $f(I)$ on $\Omega$ and use k-means clustering to partition $\Omega$ (or a subset of $\Omega$) into $k$ clusters. We use the k-d tree algorithm to recursively build a hierarchy of such clusters, specifying feature functions $f_1(I),\ldots f_n(I)$. For the first $n-1$ levels, tree nodes at level $i$ subdivide sets of images into a pre-specified number ($k_i$) of clusters, and the $n$-th level is flat.
\par As a specific example, $\Omega$ might typically contain a few thousand images. Then, choosing  $f_1=f_{pose}(I)$, $f_2=f_{lighting}(I)$, $f_3=f_{expression}(I)$, $k_1=9,$ and $k_2=3$, we would build the cluster hierarchy as follows: first, we extract $f_{pose}(I)$ for all $I\in\Omega$, and split $\Omega$ into $k_1=9$ clusters based on $f_{pose}(I)$. Within each of the nine pose clusters, we then further split the images into $k_2=3$ subclusters based on $f_{lighting}(I)$. The resulting k-d tree is thus defined by (and stored as) nine cluster centers in the pose feature space, 27 cluster centers in the lighting feature space, and a flat list of expression features $f_{expression}(I)$ in each of the 27 lighting clusters.
\par After building the k-d tree, we can then efficiently retrieve a subset of images by specifying an input image $I_{query}$, evaluating $f_1(I_{query}),\ldots$ $f_n(I_{query})$, and specifying the number of desired matches at each level of the k-d tree: $q_1,\ldots,q_n$. For example, a query with $q_1=2$, $q_2=3,$ $q_3=1$ would select $q_1=2$ best matches out of the nine pose feature cluster centers at the first level; then, within each of those 2 pose cluster matches, the query would select all $q_2=3$ lighting feature clusters at the second level (to provide good variation in lighting). Finally, within each of those six image partitions, we'd then pick the closest expression match ($q_3=1$), resulting in the retrevial of six images.
\par Figure \ref{fig:ffmatch} shows some results across different datasets, with k-d trees of depth 3, $f_1,f_2,f_3=$pose, lighting, expression, $k_1,k_2=9,3$, and $q_1,q_2,q_3=2,3,1$. Our perceptual features were designed to be as agnostic to facial appearance as possible so that given a hierarchical cluster of images of person A, one can query using images of person B.
\figcaption{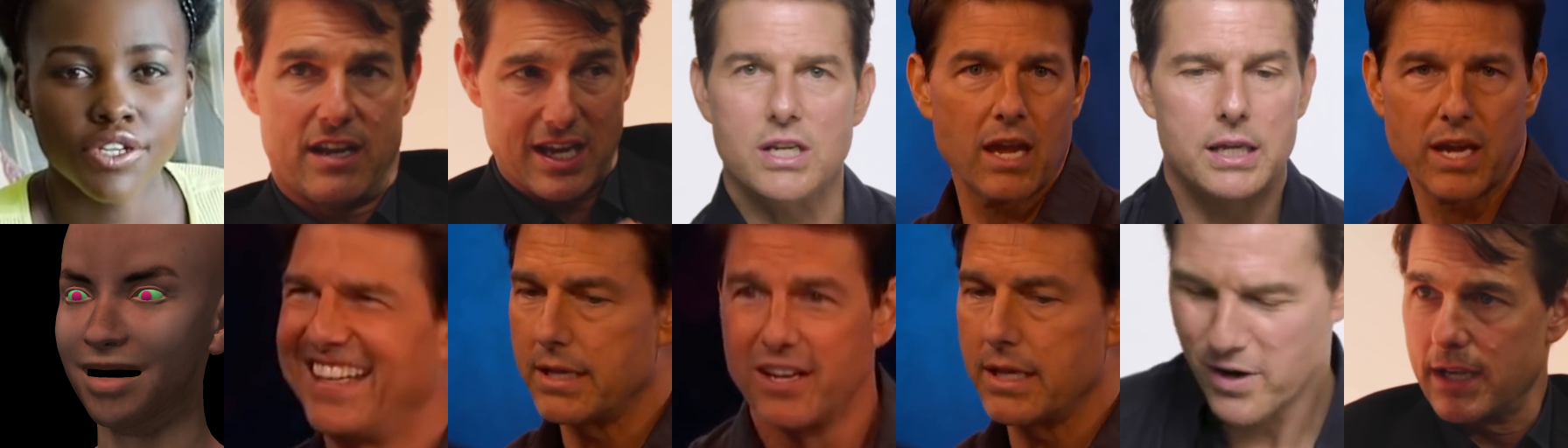}{Query image on the left, followed by 6 matches from a Tom Cruise dataset, optimized to find nearby pose and expression while maintaining variance in lighting (two examples are shown). The Tom Cruise images were gathered from four separate youtube interviews.}{fig:ffmatch}
\section{Bridging the Domain Gap}
Typically, when using synthetic data, one would strive to make the images as realistic-looking as possible in order to minimize the domain gap between the synthetically generated and the in-the-wild images. In contrast, we use a minimally complex rendering pipeline (diffuse shading, simplistic textures, etc.) and bridge the domain gap with unsupervised conditional autoencoders as used in deepfake technology. Using only a small set of curated training data for each subject, we utilize deepfake technology to correlate and transform between synthetic and in-the-wild images. Serendipitously, this allows us to aggressively ``annotate" synthetic renders, and subsequently transport that annotation to in-the-wild images (see e.g. Section 4.1).
\par For the deepfake component of our pipeline, we use a single encoder ($E$) and fully decoupled dual decoder ($D_A,D_B$) architecture \cite{perov2021deepfacelab}. During training, aligned crops of both synthetically-generated and in-the-wild images are fed into the joint encoder $E$, before being passed through the two separate decoders $D_A,D_B$. An image reconstruction loss
$$f_{recon}=\nrm{D_A(E(I_A))-I_A}+\nrm{D_B(E(I_B))-I_B}$$
and discriminators $G_{image},G_{latent}$ paired with asymmetric GAN loss
\eqnsplit{
f_{image}&=\nrm{G_{image}(D_A(E(I_A)))}+\nrm{\vec{1}-G_{image}(I_A)}\\
f_{latent}&=\nrm{G_{latent}(E(I_B))}+\nrm{\vec{1}-G_{latent}(E(I_A))}
}
serve as the main penalty constraints of the decoder outputs during training. As is typical, $G_{image}$ attempts to discern whether the image comes from $\Omega_A$ or the decoder $D_A$ (minimizing $f_{image}$), whereas $D_A$ attempts to fool the discriminator (maximizing $f_{image}$). $G_{latent}$ attempts to discern whether the latent image encoding $E(I)$ was generated from an image $I\in\Omega_A$ or $I\in\Omega_B$ (minimizing $f_{latent}$), whereas $E$ attempts to fool the discriminator (maximizing $f_{latent}$). 
The asymmetric GAN loss $f_{image}$ could be easily reflected and made symmetric by incorporating a matching set of discriminators and losses, but in practice, we care specifically about the results of one specific decoder rather than both (e.g. the `real' decoder for appearance capture in Section 5, and the `synthetic' decoder for motion capture in Sections 8 and 9.) 
The final results of unsupervised training on two sets of facial images $\Omega_A=\set{I_A}$ and $\Omega_B=\set{I_B}$ are a learned encoder $E$ that encodes and correlates $\Omega_A$ and $\Omega_B$ into a joint embedding space, and two separate decoders $D_A,D_B$ that can decode embedded vectors into images that seemingly belong to either $\Omega_A$ or $\Omega_B$ (see Figure \ref{fig:df}). 
\figcaption{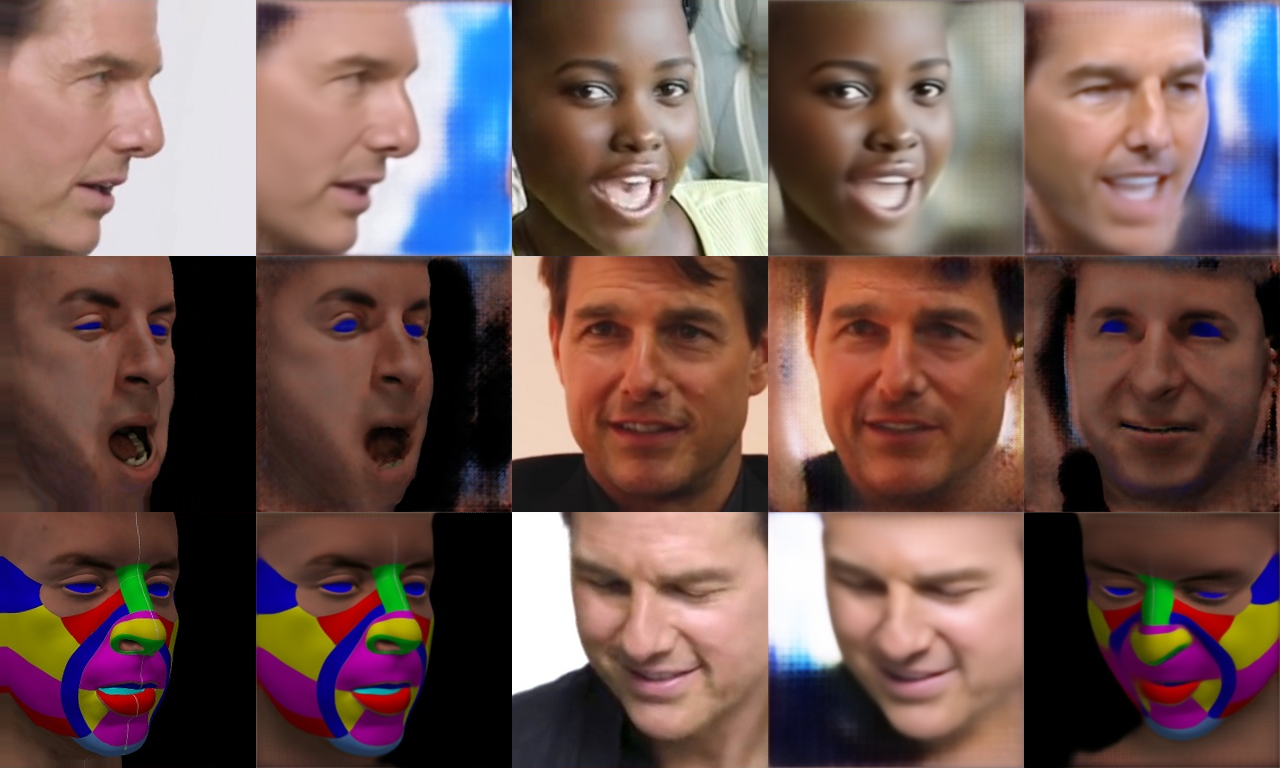}{Column 1 shows three different examples of an image $I_A$ from a set $\Omega_A$ (Tom Cruise data, synthetic renders, synthetic segmentation renders). Column 2 shows $D_A(E(I_A))$. Column 3 and 4 show similar results for $I_B$ and $D_B(E(I_B))$ (Lupita Nyong'o, Tom Cruise, Tom Cruise). Finally, column 5 shows $D_A(E(I_B))$.}{fig:df}

\subsection{2D Motion Tracking} 
The goal of 2D image based tracking is to draw feature points (and/or curves) on images in order to annotate motion. The most successful and robust landmark detection methods typically use CNN based approaches, even though CNNs work with areas of pixels, not codimension one features. The common method of addressing this discrepancy is to leverage area based heatmaps by detecting local extrema to determine tracked points. 
By drawing on ideas from levelset methods (see e.g. \cite{osher2001level}), we instead embed codimension one points and codimension two curves as the boundaries of tracked areas. It is well known that one can track levelsets of functions far more accurately and robustly than extrema. 
\par To enable this approach, we segment the 3-dimensional face surface into regions, with boundaries that are a superset of the codimension one and codimension two features we wish to track. Then, given a set of in-the-wild images $\Omega_R$, we construct a synthetically rendered dataset $\Omega_S$ using the segmented 3D face. Once the deepfake network is properly trained, any image $I_R\in \Omega_R$ can be passed into the encoder and decoded as $D_S(E(I_R)).$ Identifying the boundaries between regions on $D_S(E(I_R)$ gives the desired codimension 1 and 2 features that we wish to track. (See Figure \ref{fig:dfseg}.)
\figcaption{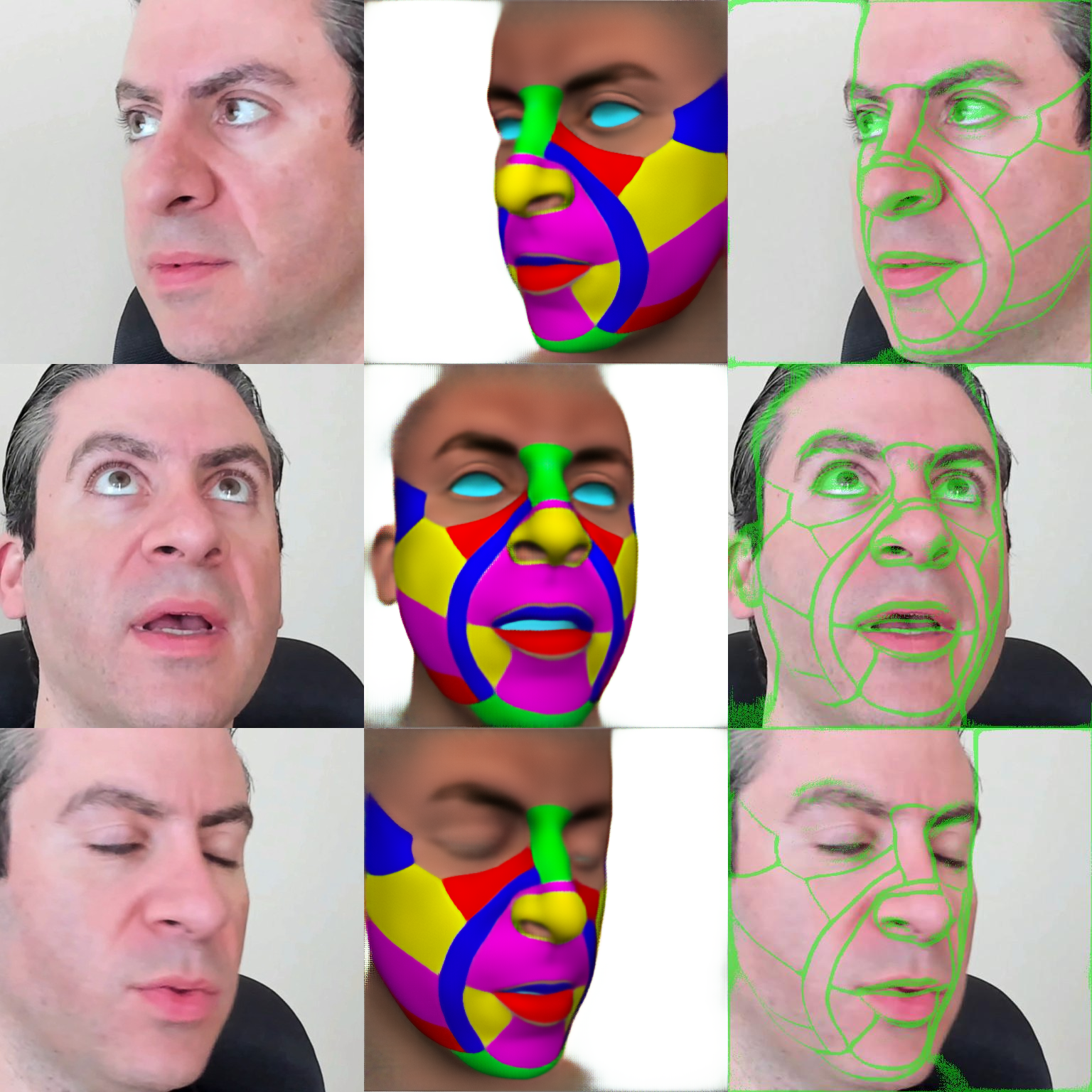}{We train a deepfake network on webcam footage $\Omega_R$ and synthetic segmented renders $\Omega_S$. From left to right: real image $I_R\in\Omega_R$,  deepfake results $D_S(E(I_R)$, and the boundary contours of $D_S(E(I_R))$ overlaid onto $I_R$.}{fig:dfseg}
\section{Appearance Capture}
Revolving (theater) stages, colloquially known as turntables, are an effective way of systematically evaluating (and iterating upon) the design of 3D models. Examples can be seen in advertising, car shows, art displays, sculpts for practical special effects, etc. 
Synthetic renders of steadily rotating computer-generated objects give a holistic view of an object's geometry and texture, allowing one to avoid 3D geometry that only looks good from a subset of viewpoints. In addition, multi-view stereo synthesis of geometry is well-established in computer vision, and core to many classical methods (visual hull, structure from motion, etc.). Multi-view stereo approaches for facial capture traditionally depend on having fine-grained control over the capture environment, making such approaches impractical for democratization. In contrast, when working with synthetic renders instead of real-world capture data, one has full control over both geometry and camera, which we can leverage to create a dense and precise multi-view pipeline. Our appearance capture pipeline is thus centered around turntable based evaluation and synthesis of 3D geometry.
\subsection{Geometry Estimation} To bootstrap our approach, we utilize an existing single-view based method \cite{sanyal2019learning} to get an inital estimate of the geometry. Although such approaches allow for the democratization of appearance capture, they suffer significantly from dataset bias because single-view reconstruction is a highly underconstrained problem. Thus, we prefer a personalized approach when possible (e.g. a classical high-end scanner, a multi-view neural network, depth range, etc.). More specifically, we recommend starting with an estimate of the geometry and subsequently refining it using inverse rendering. We have had particular success using the segmented face texture from Section 4.1 (see Section 8 for more details). Regardless, any method suited for democratization will incur errors in the reconstructed geometry; thus, it is necessary to utilize a texture acquisition method that is able to cross the domain gap between imperfect geometry and real-world images.
\subsection{Data Curation} Given estimated geometry, we generate a set of 200 synthetically rendered images with varying pitch and yaw (see the discussion of pose in Section 3).
Additionally, given in-the-wild footage of the subject under consideration, we build a k-d tree as discussed in Section 3. Then, for each of the 200 synthetic turntable renders, we use the k-d tree to find a few best matches in pose and expression. The end result is about 200-300 unique images from the in-the-wild footage, all with a relatively fixed expression (matching the synthetic geometry) across a wide range of views. This data curation step enables our method to work on input data ranging from five minute youtube clips of people talking and gesticulating to 30 second shaky self-recorded videos taken on handheld smartphones.  See Figure \ref{fig:ffturn}.
\figcaption{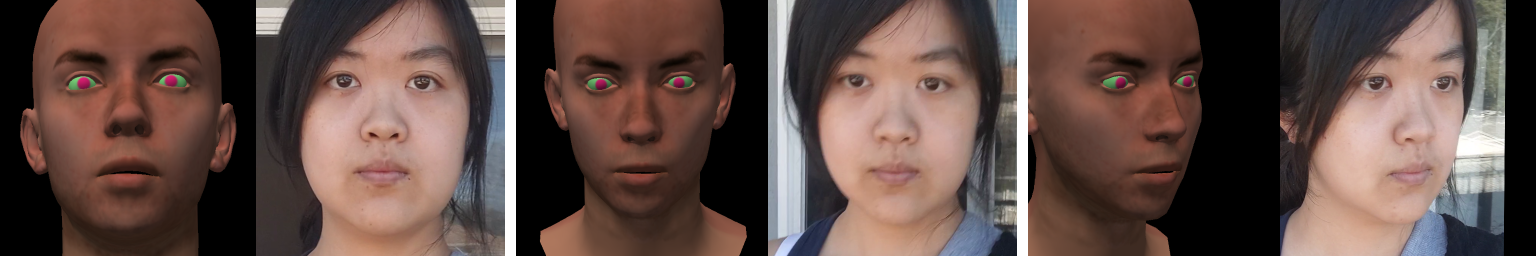}{Example matches to synthetic renders found in real-world video footage using hierarchical clustering. Note that instead of having a fixed pitch and varying yaw as is standard for turntables, we found varying the pitch of the face to be quite useful as well.}{fig:ffturn}
\subsection{Deepfake Training} Given the 200 synthetic renders and the similarly-sized pruned in-the-wild dataset, we train a deepfake network (see Section 4) with input and output sizes of $384\times 384$ pixels. With these small and heavily correlated datasets, it takes less than an hour for the network model to converge using a single NVIDIA RTX GPU. By utilizing a slightly-pretrained model as a warm start, one can further cut this down to under half an hour.
\subsection{Texture Acquisition: Face} 
We take a subset of 20 synthetic turntable renders, and for each render $I$, we generate a pixel-by-pixel rasterization $U(I)$ of the 3D model's UV coordinates as well as a deepfake result $D(I)=D_R(E(I))$. See Figure \ref{fig:dfuv}.
\figcaption{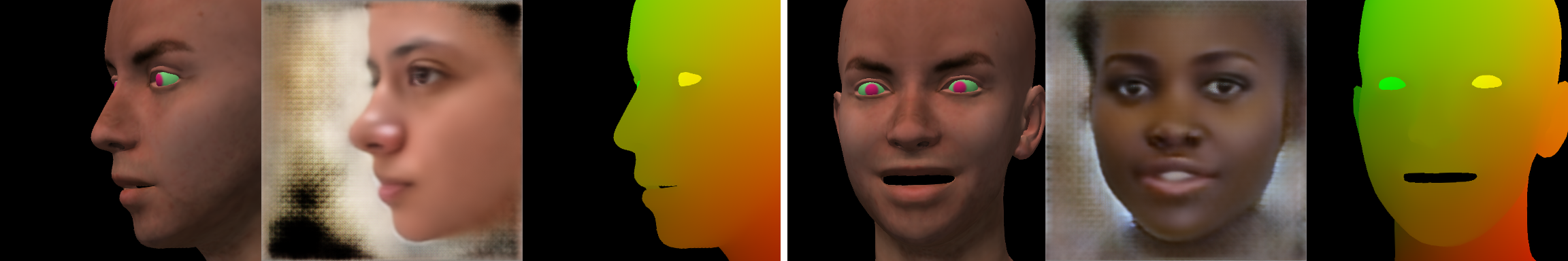}{Left to right (two examples shown): Given a synthetic turntable render $I$, the deepfake result $D(I)$ is well aligned to it, allowing us to generate correspondences between the deepfake pixels and the rasterized UV coordinates $U(I)$ of the model.}{fig:dfuv}
Motivated by photon gathering \cite{jensen2001realistic}, for each valid pixel, we copy its color $\vec{c}\in D(I)$ into the UV space as a particle sample at its corresponding location $\vec{p}\in U(I)$. After consolidating particle samples from all 20 deepfake images into a single photon map, we use photon gathering to generate the final texture image $T$. As is typical in photon mapping, the size of the 2D disc around each texel coordinate is increased or decreased to collect a fixed number of particle samples, and the colors of the collected samples are subsequently averaged to obtain the final texel color. We use an inverse-distance weighted average to aggregate the samples. Additionally, each texel is assigned a confidence score inversely proportional to the disc radius. See Figure \ref{fig:texface}.

\figcaption{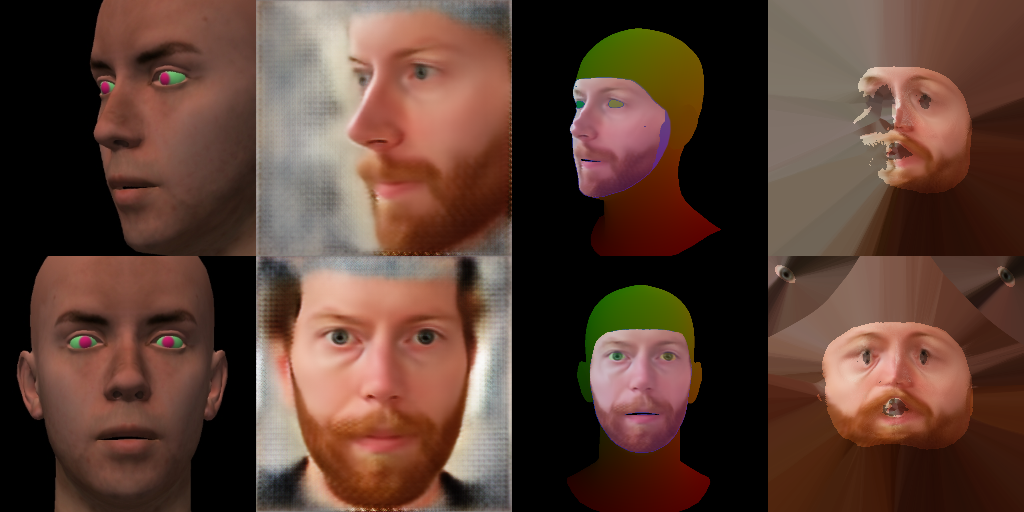}{Left to right: Synthetic render, deepfake, high confidence deepfake pixels overlaid onto UV coordinates, and the texture map generated from correspondences between deepfake pixel and UV coordinate pairs. The first row shows results generated using a single view. The second row shows cumulative results generated using multiple views (including the views shown.)}{fig:texface}
\subsection{Texture Acquisition: Head}
Deepfake networks are most robust when constrained to a tight crop of the face. In order to ascertain texture in regions where we do not have accurate deepfake pixels, we use hierarchical clustering (see Section 3) to obtain in-the-wild image matches to the synthetic turntable renders. The matches are close but not perfect, often due to slight variation in expression as well as errors in the geometry estimation step; therefore, we strongly prefer to use aligned deepfake pixels (Section 5.4) whenever possible (see Figure \ref{fig:rndfcmp}).
 For each synthetic render $I$, after finding the best in-the-wild image match, we adjust the 3D model to fit that match as closely as possible (as discussed regarding the pose feature in Section3). Then, we generate $U(I)$ and use the selected in-the-wild image in place of $D(I)$ to generate a texture map along the lines of Section 5.4. See Figure 10. Afterwards, the results are combined with the texture map from Section 5.4, using the per-texel confidence scores mentioned in that section. See Figure \ref{fig:texhead}.
\figcaption{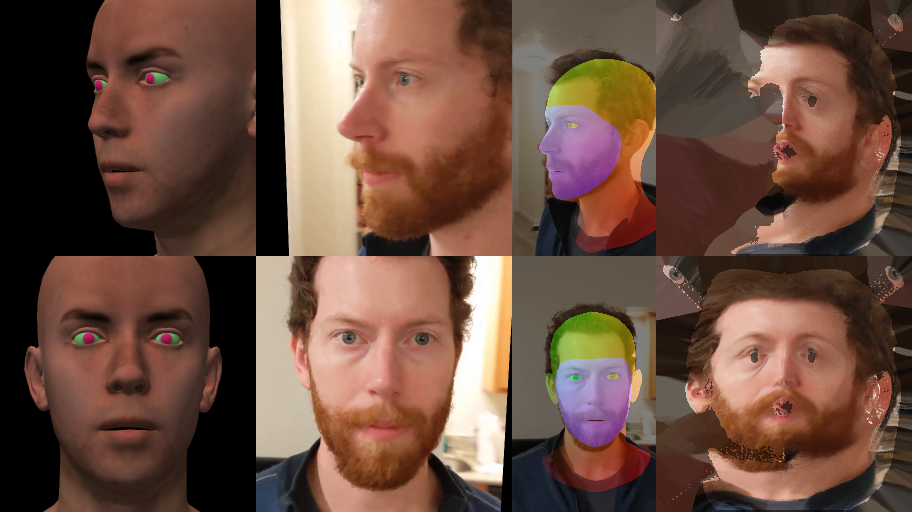}{Left to right: Synthetic render, closest dataset match from hierarchical clustering, UV coordinates of the (pose-fitted) synthetic model overlaid onto the source plate of the closest match, and the texture map generated from correspondences between source plate pixel and UV coordinate pairs.  The first row shows results generated using a single view. The second row shows cumulative results generated using multiple views (including the views shown).}{fig:texhead}
\subsection{Results} 
\begin{figure*}[h]
 \centering
  \includegraphics[width=\textwidth]{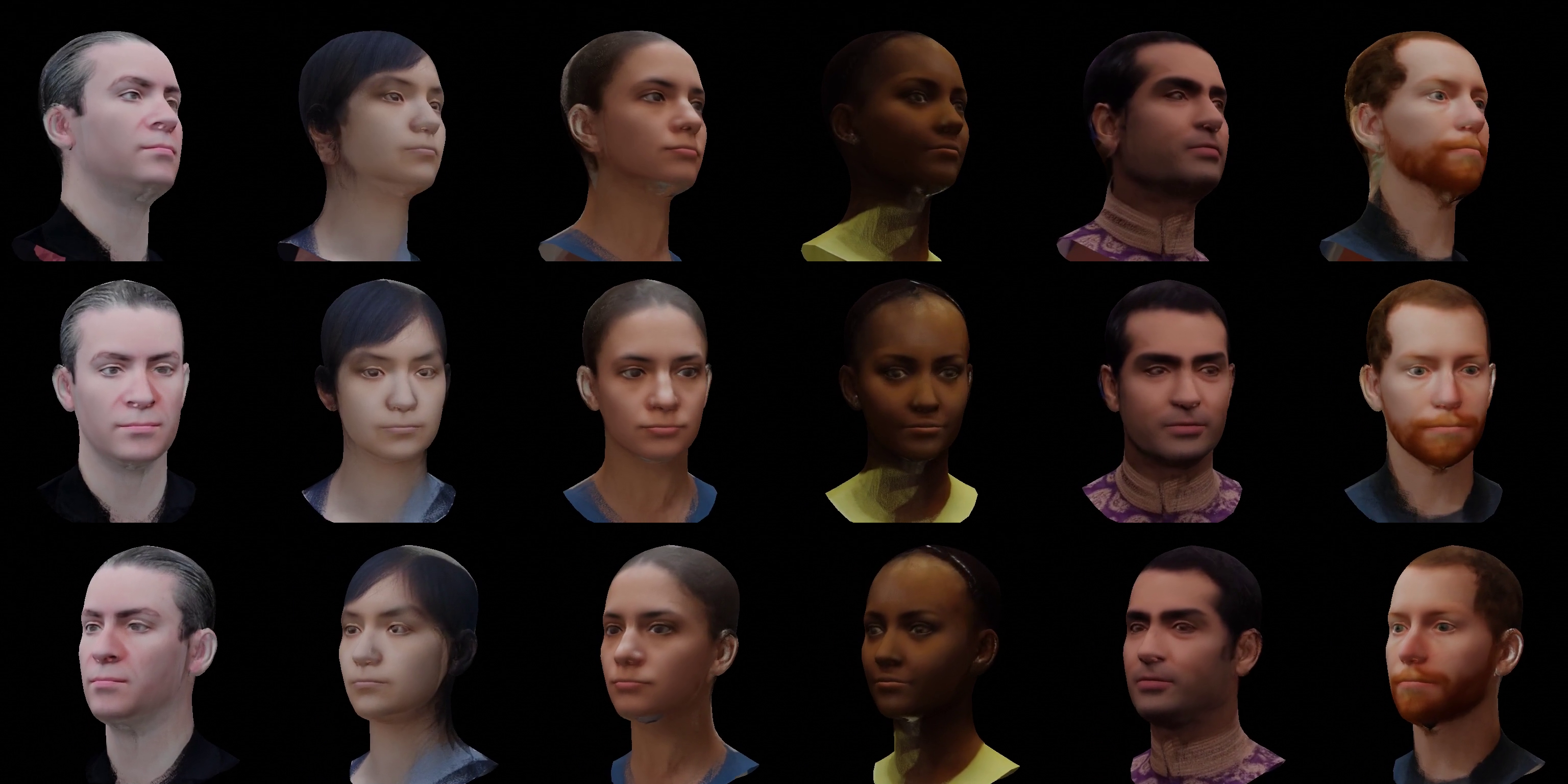}
  \caption{Results on six subjects, rendered from several views. We present the raw results without any manual cleanup in order to illustrate the potential for democratizing our approach. It is straightforward to utilize matting networks, etc. to clean up the texture maps.}
  \label{fig:texresult}
\end{figure*}
We evaluated our pipeline on data ranging from 20 seconds of video from a handheld smartphone camera to 5 minutes of video from youtube celebrity interviews. See Figure \ref{fig:texresult}.
Starting from the extraction of frame-by-frame data from raw in-the-wild video input, texture maps for each subject were generated within an hour. Compared to traditional multi-view stereo, our deepfake approach produces fewer misalignment artifacts when paired with imperfect geometry. In particular, the improvements are most striking on non-Caucasian female subjects (see Figure \ref{fig:rndfcmp}), since the monolithic network we use for predicting initial geometry struggles the most on these subjects (presumably due to the fact that they are not well-represented in the dataset used to train the network).
\par The biggest limitation of our work currently lies with the resolution constraints of deepfake networks. Since our approach does not require modifications of existing dual auto-encoder technology, we foresee that the efficacy of our results will improve as deepfake technology becomes scalable to higher resolutions. Additionally, while an automated pipeline that can run in under an hour is already tractable for democratization, we foresee that more aggressive optimizations would be required in order to scale the approach to millions of casual users on their home devices.
\figcaption{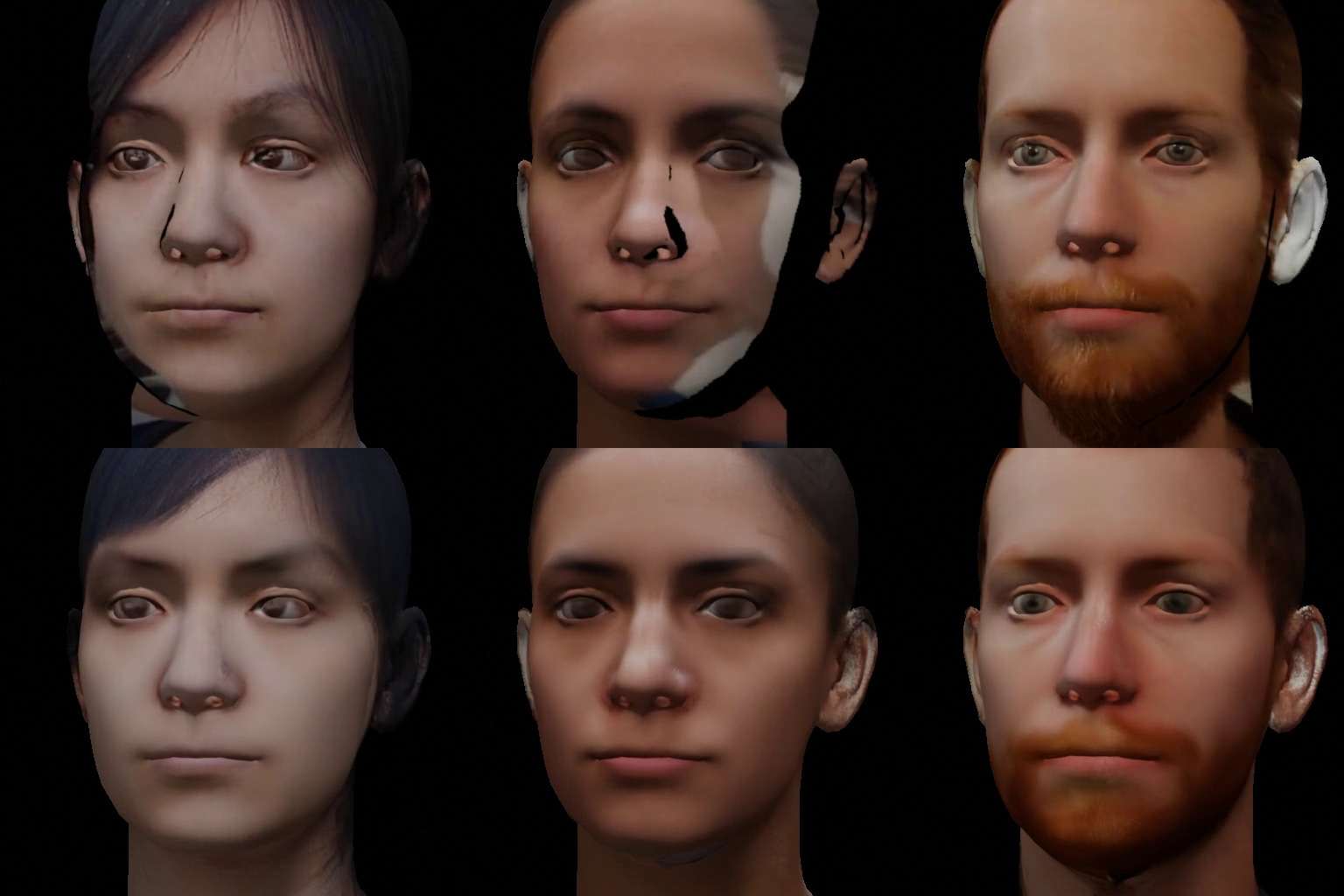}{The top row shows a texture generated by splatting in-the-wild imagery onto imperfect predicted geometry, with our texture results on the bottom row. Note that the misalignment in the top row is particularly troublesome around the eyes and nose.}{fig:rndfcmp}
\par An interesting line of future research would be to build texture maps that can be used with more sophisticated lighting and shading, such as separated specular, diffuse, and subsurface-scattering texture maps. The turntable synthetic renders would then also need to incorporate lighting into consideration, necessitating a virtual lightstage that achieves similar results to \cite{texturemaps} through deepfake technology.
\section{Motion Capture}
We can transfer any animation rig to the textured geometry resulting from Section 5, e.g. via a volumetric morph defined by surface-curve based boundary constraints \cite{cong2015fully}. In particular, we begin by transferring the Metahuman \cite{metahuman2021} joint based rig to our Section 5 results. See Figure \ref{fig:texanim}.
\figcaption{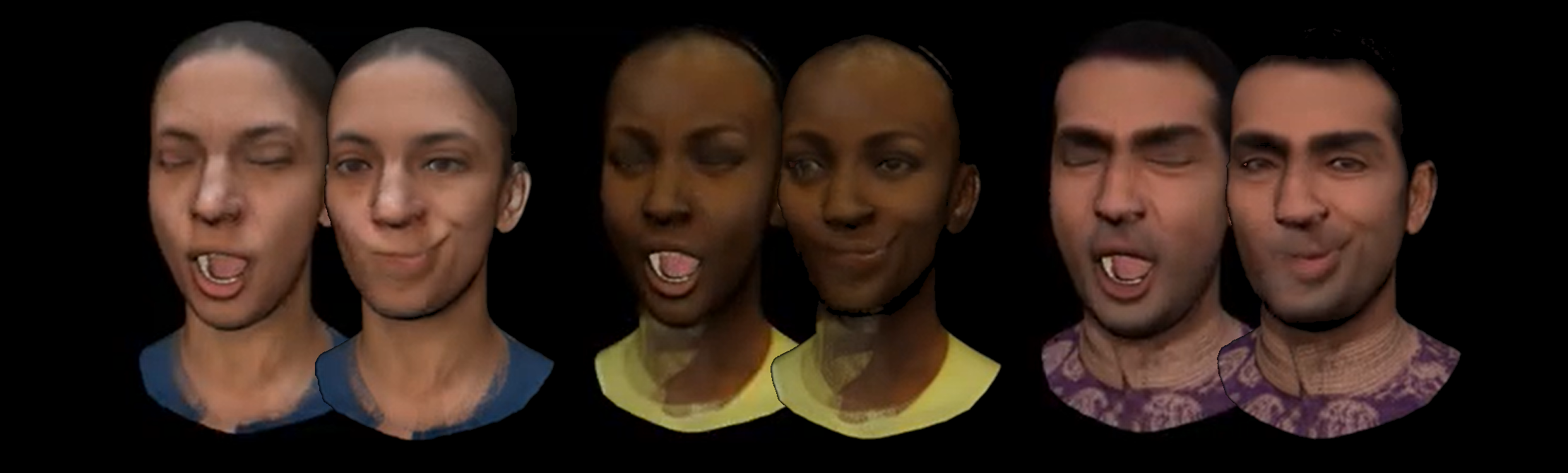}{After morphing the metahuman rig to our textured geometry, we can evaluate our appearance capture results on a range of (retargeted) artist directed animations.}{fig:texanim}
The Metahuman rig is composed of 167 user controls, 253 shapes, and 683 shape correctives on a surface mesh with roughly 24,000 vertices. For the purposes of motion capture, we simplify this rig by converting it to a linear blendshape rig with jaw joint skinning, discarding the correctives. More in-depth discussions of expression rigs can be found in \cite{egger20203d}.
\par For each shape $S_i$ in the original rig, we split it into deformations caused by the jaw rotation and translation ($R_i$ and $T_i$) and the "de-skinned" offset $D_i$ from the jaw-induced deformation. Given the per-vertex jaw-skinning coefficient tensor $M$ and neutral shape $N$, each shape can then be described as \eqnsplit{S_i&=M\Big(R_i(N+D_i)+T_i\Big)+(1-M)\Big(N+D_i\Big)\\&=\Big(MR_i+(1-M)I\Big)(N+D_i)+MT_i}
where the entries of $M$ vary between 0 and 1 (equal to 0 for vertices only affected by the skull, and 1 for vertices only affected by the jaw). This rig is parametrized by $\vec{w}\in[0,1]^{253}$ via
{\small\eqnsplit{S(\vec{w})=\Big(MR(\vec{w})+(1-M)I\Big)(N+\sum w_iD_i)+MT(\vec{w})}}
where the jaw rotation matrix $R(\vec{w})$ is generated via Euler angle interpolation of the $R_i$, and $T(\vec{w})=\sum w_iT_i$. The nonrigid deformations $S(\vec{w})$ combined with the rigid transformation of the skull $P_{rigid}$ fully defines the range of motion we wish to capture. See Figure \ref{fig:rig}.
Motion tracking is thus defined as extracting the pose $P_{rigid}$ and control rig parameters $\vec{w}$ from images.
\figcaption{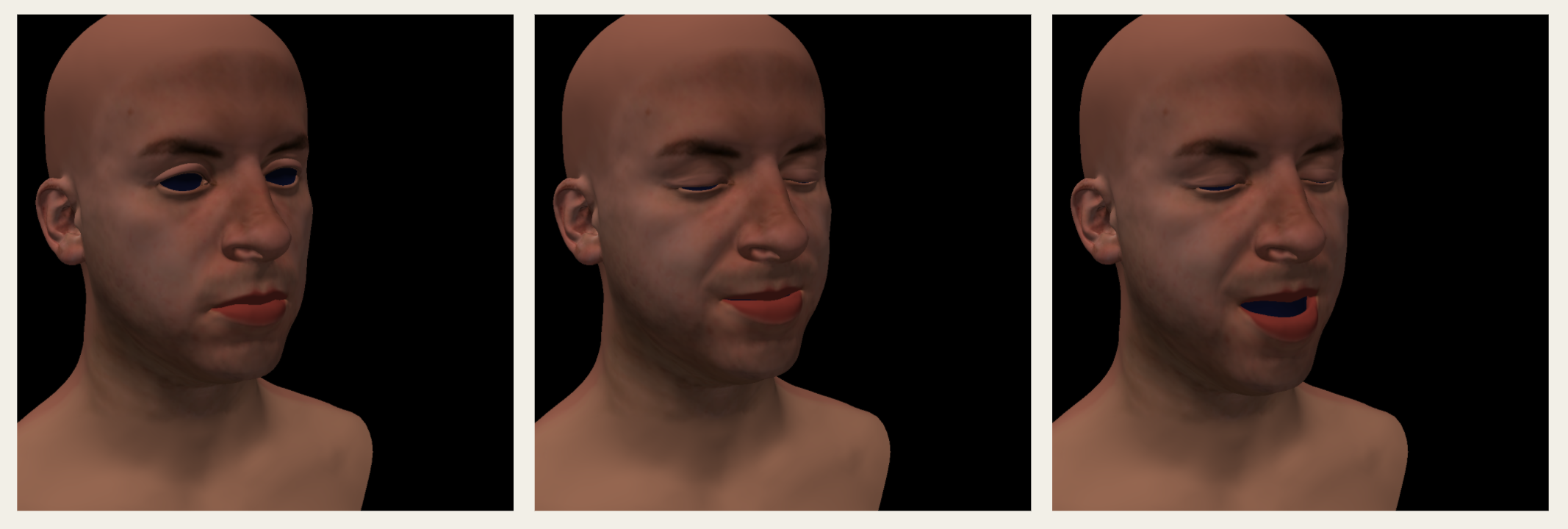}{From left to right, the neutral face is shown in a rigid pose $P_{rigid}N$ (rotated to the left), with linear deformations $P_{rigid}(N+\sum w_iD_i)$, and with both linear deformations and jaw skinning $P_{rigid}S(\vec{w})$.}{fig:rig}
\par In Section 7, we discuss data curation with a specific focus on generating datasets to train deepfakes that can be used to facilitate motion-tracking. In sections 8 and 9, we present two approaches (inverse rendering and learned regression) for deepfake-guided tracking of personalized models.

\section{Motion-Tracking Deepfakes}
In order to train deepfake models for turntable synthesis, we pruned in-the-wild data to match the synthetically-generated dataset (see Section 3). Conversely, here we augment the synthetic dataset to match the in-the-wild data we wish to track. Ideally, the synthetic images would be generated from a control rig parameter distribution that closely matches the motion present in the in-the-wild data, but acquiring such a distribution is a ``chicken-and-egg" problem as one cannot generate such synthetic data until the in-the-wild footage has already been properly tracked. The problem is doubly difficult when working with a high fidelity rig (as opposed to a simpler PCA rig), since the high-dimensionality of such a rig provides for increased expressivity at the cost of a large nonrealistic-expression subspace.
\par Contemporaneously with our work, \cite{moser2021semi} similarly used deepfake technology for motion tracking, addressing the data distribution problem by simultaneously generating animation rigs and control parameter samples via dense 3D motion capture (utilizing head mounted cameras and a predefined corpus); however, this approach does not seem to be intended for democratization.
With personalized rigs in mind, we propose a dataset generation method that is minimally dependent on the underlying rig and does not require any (high-end) capture setup. 
\par For the sake of exposition, we illustrate our pipeline via a typical example: first, we generate 4,000 random samples by uniformly sampling each control parameter between 0 and 1 subject to some sparsity constraints for each subregion of the face (e.g. no more than five lip shapes would be activated at a time). Each sample is then paired with a rigid pose sampled from a truncated Gaussian distribution (pitch $\in[-10^\circ,10^\circ]$ and yaw $\in[-80^\circ,80^\circ]$). Additionally, an in-the-wild dataset is constructed from 3 separate youtube clips of a specific actor. Given both the synthetic and the in-the-wild datasets, a deepfake network is trained to obtain a joint latent-space embedding.
Section 7.1 describes how we use latent space analysis to remove part (about 20 percent) of the synthetic dataset, and Section 7.2 describes how we generate solver-bootstrapped synthetic data to augment the synthetic dataset (to obtain approximately 5,000 images in total).
\subsection{Dataset Contraction} The increased expressivity of high-dimensional rigs also (unfortunately) leads to unnatural geometric deformations for a wide range of control parameters. Naively sampling the control parameter space results in many expressions that would never be present in in-the-wild data. Thus, we prune randomly generated samples via latent space analysis of trained deepfakes.
\figcaption{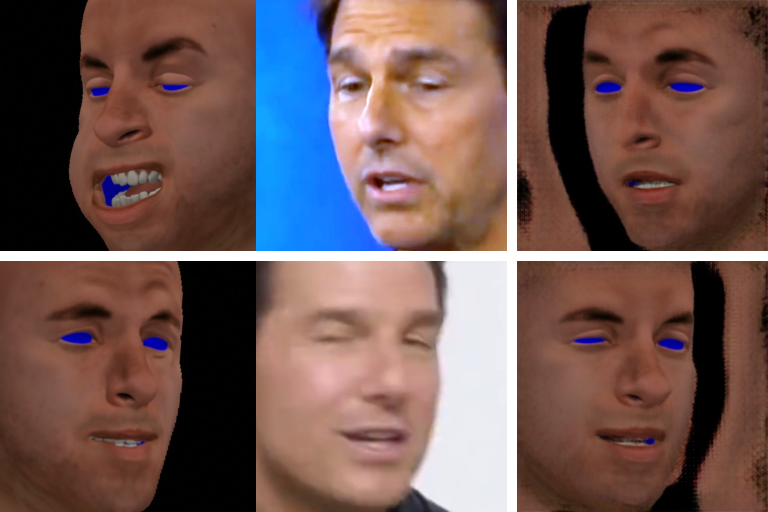}{Synthetic renders $I_S$ (left column), and the closest in-the-wild images $I_R$ (middle column) as measured in the latent space.  The top row shows an example of `out of distribution' synthetic data, contrasted with the bottom row which shows `in distribution' synthetic data. Comparing $D_S(E(I_R))$ (right column) to $I_S$ (left column) gives one a sense of the distance between $E(I_S)$ and $E(I_R)$.
 }{fig:outliersynth}
To accomplish this, we first train an unsupervised deepfake model using the randomly-generated synthetic and in-the-wild datasets. Then, taking advantage of the unsupervised clustering that occurs as a byproduct of training the deepfake network, we identify (and delete) \textit{out-of-distribution} synthetic data by finding the outliers (as compared to in-the-wild data) in the deepfake network's latent space. See Figure \ref{fig:outliersynth}.
\subsection{Dataset Expansion} Similar to our dataset contraction approach, we perform dataset expansion by (instead) finding outliers in the real data (as compared to synthetic data), characterizing them as \textit{gaps} in the synthetic dataset, and subsequently utilizing existing motion capture solvers to bootstrap \textit{gap}-filling synthetic data. See Figure \ref{fig:outlierreal}. Extracting control parameters from real images is precisely the problem we wish to solve, so we are again faced with a chicken-and-egg problem.
Solely relying on existing solvers to expand the synthetic dataset is not only over-reliant on the accuracy of the solvers, but also causes the network to overfit to and mimic these solvers (as opposed to being trained to match the in-the-wild footage in an optimal way). To avoid this, we only use off-the-shelf solvers to bootstrap the dataset expansion, and subsequently augment bootstrapped data with random jittering and interpolation. More specifically, given a control parameter sample $\vec{w}$ generated from an existing solver, we add per-control random noise to generate new samples $D\vec{w}$, $D$ being a diagonal matrix with diagonal entries $d_{i}\in [0.8,1.2]$. In addition, given two control parameter samples $\vec{w}_1$ and $\vec{w}_2$, we  interpolate between them via  $\alpha\vec{w}_1+(1-\alpha)\vec{w}_2$ with $\alpha\in(0,1)$. Of course, any newly added samples can be examined for outliers (and pruned if necessary) using the approach in Section 7.1.

\figcaption{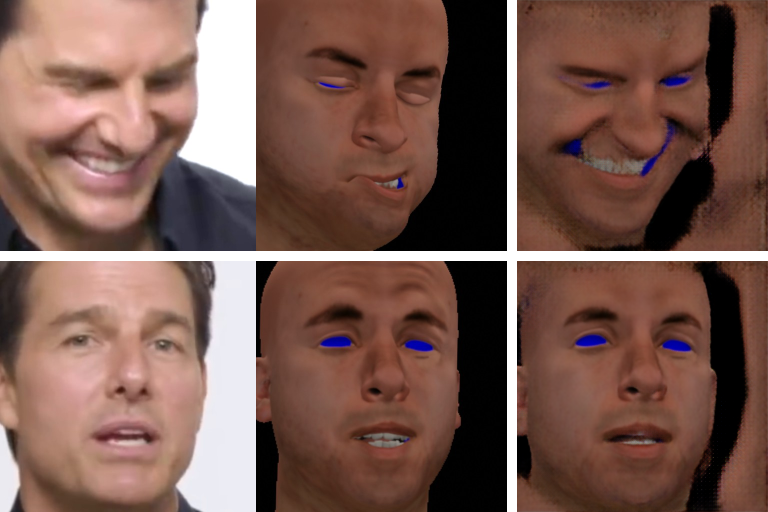}{In-the-wild images $I_R$ (left column) and the closest synthetic renders $I_S$ (middle column) as measured in the latent space. The top row shows `out of distribution' in-the-wild data, contrasted with the bottom row which shows `in distribution' in-the-wild data.  Comparing $D_S(E(I_R))$ (right column) to $I_S$ (middle column) gives one a sense of the distance between $E(I_R)$ and $E(I_S)$. }{fig:outlierreal}
\section{Inverse rendering}
Inverse rendering differentiates through the rendering pipeline in order to optimize parameters for geometry, material properties, lighting, etc. The restriction that the renderer has to be fully differentiable limits its visual fidelity, and thus its ability to match photorealistic in-the-wild images. Most practitioners are well-aware of such limitations. Here, we present a promising approach for using inverse rendering on in-the-wild images, by utilizing segmentation deepfakes (as described in Section 4.1). Inferencing a segmented texture from an in-the-wild image (using a trained deepfake) allows one to more robustly determine parameters via an inverse rendering approach.
\figcaption{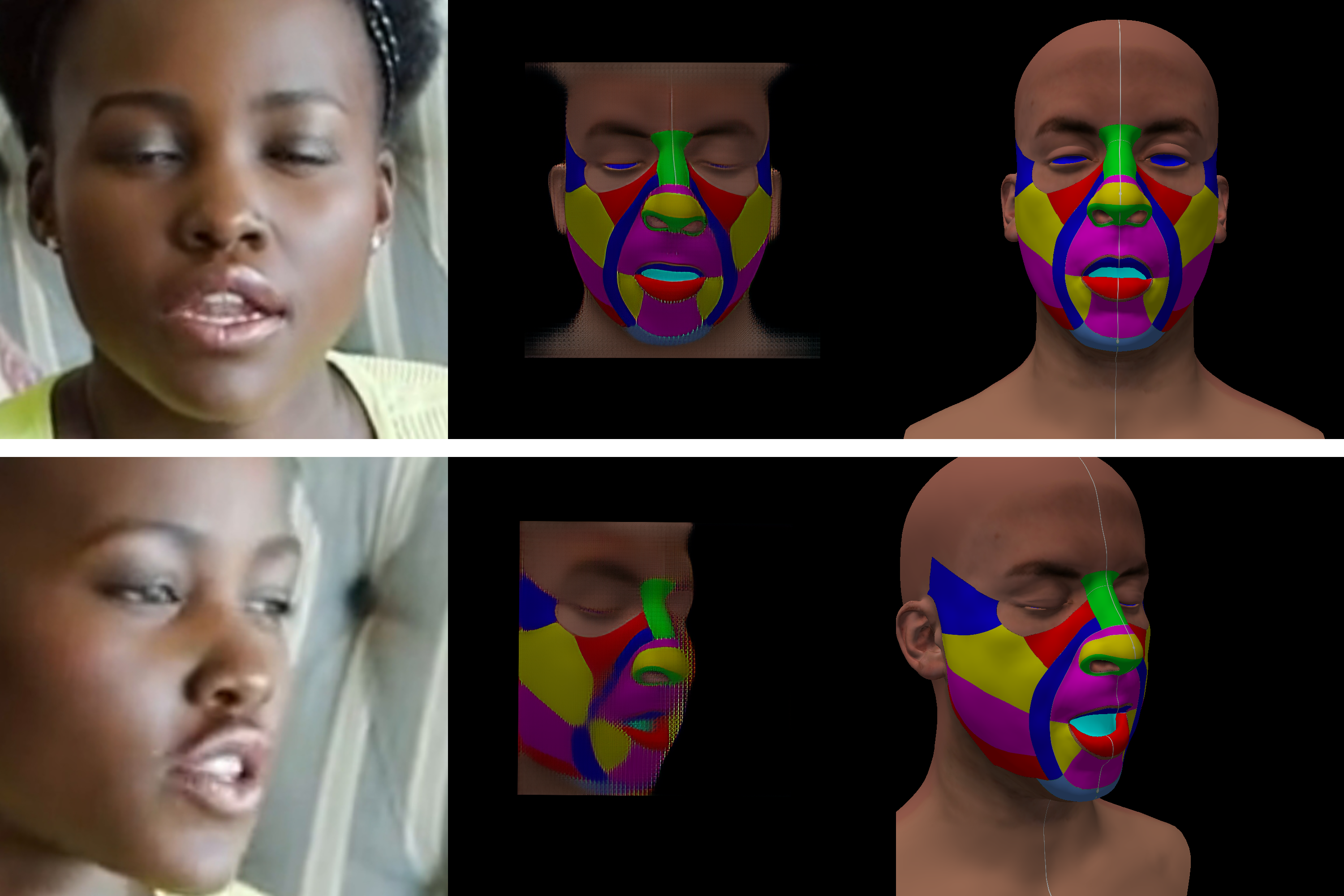}{Two examples of inverse rendering solves. From left to right: input frame $I_R$, $D_S(E(I_R))$ rigidly transformed to the blendshape rig's image coordinate frame, and a rendering of the blendshape rig with solved parameters.}{fig:irsolve}
\par To do this, we first implement our blendshape rig as a fully-differentiable pytorch3d \cite{ravi2020pytorch3d} module. Given a fixed segmentation texture $T_{seg}$ (as opposed to a photorealistic face texture $T_{real}$),  $$F_{seg}(p,y,\vec{w})=F(P_{rigid}(p,y)S(\vec{w});T_{seg})$$ is a differentiable image generating function parametrized by $(p,y,\vec{w})$. Then, given an in-the-wild dataset, we use the method described in Section 7 to generate a synthetic segmentation dataset ($\Omega_S$) that matches the statistical distribution of the in-the-wild data ($\Omega_R$);
afterwards, a deepfake model is trained between the two datasets.
\par Given an in-the-wild image $I_R$, we solve  $${\arg\min}_{p,y,\vec{w}}\nrm{F_{seg}(p,y,\vec{w}) - D_S(E(I_R))}.$$ using a pixel-wise L2 distance and block coordinate descent, i.e. each epoch first considers the rigid pose, then the subset of control parameters active for jaw-skinning, finally followed by the rest of the control parameters. See Figures \ref{fig:irsolve} and \ref{fig:irblocksolve}.
\figcaption{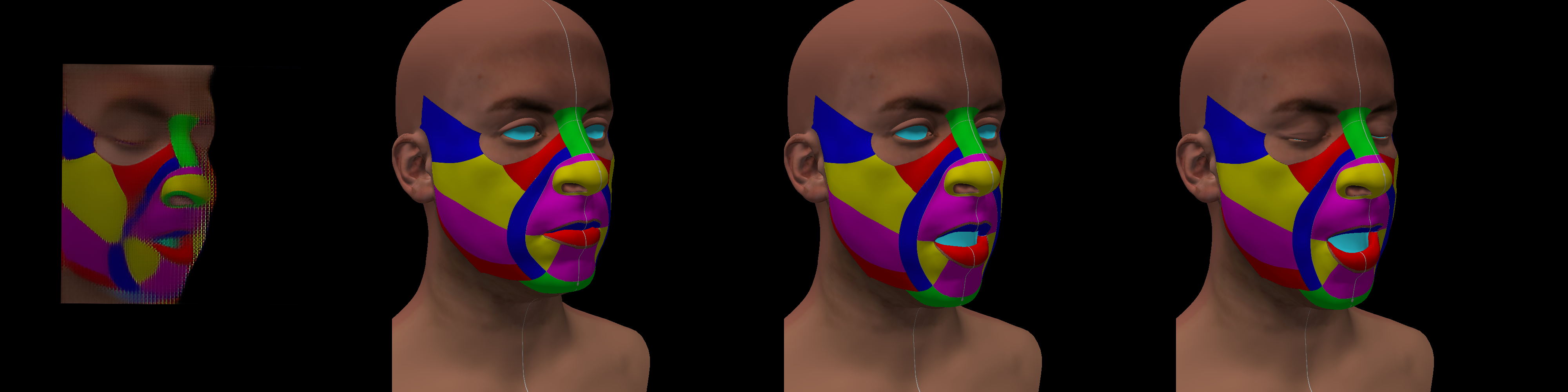}{A breakdown of the block gradient descent algorithm. Given the deepfake result on the left, we iteratively solve for pose, jaw, and expression parameters (from left to right).}{fig:irblocksolve}

\par For the sake of comparison, we experimented with both realistic and segmented textures, training separate deepfake models to inference either the segmented or realistic texture onto in-the-wild images. We observed lower sensitivity to the initial state and faster convergence when using the segmented (as opposed to the realistic) texture. See Figure \ref{fig:segrealcmp1}. In particular, one can typically use the front-facing neutral expression as a robust starting point for every frame; however, starting from an initially computed pose (as described in Section 3) further increases robustness. Note that subspace analysis (described in Section 7.1) can also be used to provide a robust initial guess.
\figcaption{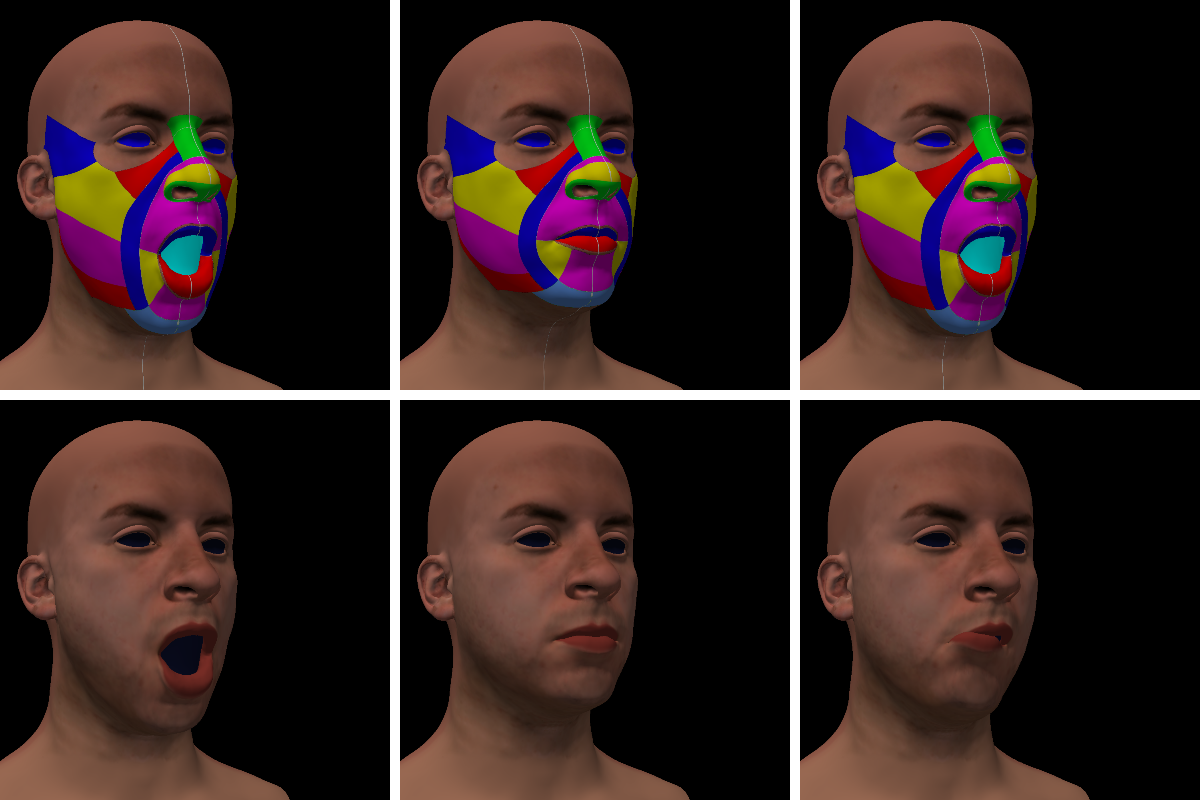}{A simple test comparing our use of $F_{seg}(p,y,\vec{w})$ (top row) with the use of a more photorealistic $F_{real}(p,y,\vec{w})=F(P_{rigid}(p,y)S(\vec{w});T_{real})$ (bottom row) for the inverse rendering. Given the same target pose and expression  $p^*,y^*,\vec{w}^*$ (first column), the inverse renderer using $T_{seg}$ starting from the correct rigid pose (column 2) converges as expected (column 3); replacing $T_{seg}$ with $T_{real}$ gives poor results.}{fig:segrealcmp1}
\figcaption{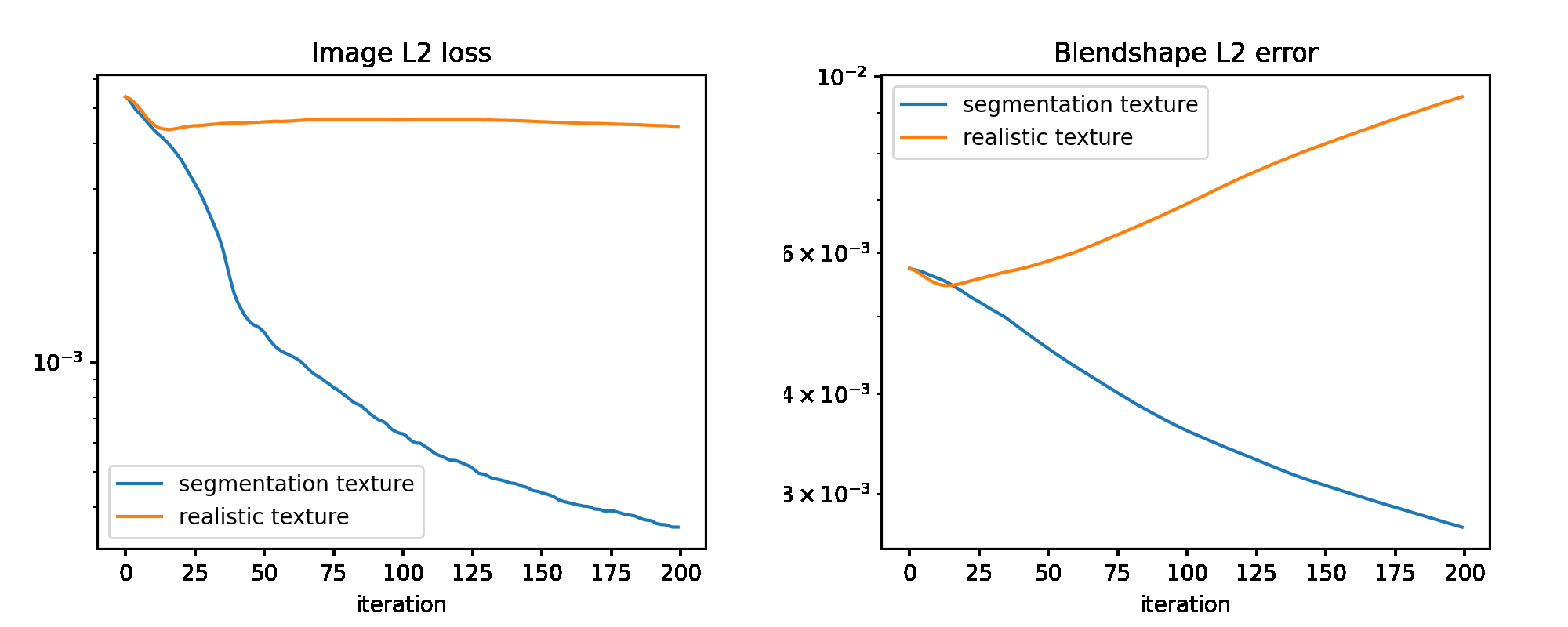}{Plots showing the L2 image loss (which our inverse renderer optimizes over), as well as the L2 error on $\vec{w}.$ The blue line corresponds to the first row of Figure \ref{fig:segrealcmp1} using $F_{seg}$, and the orange line corresponds to row 2 using $F_{real}$.}{fig:segrealcmp2}
\par In our experiments, we were able to solve for full in-the-wild sequences without manual supervision at 10-30 seconds per frame. Moreover, the ability to use the neutral expression as an initial guess for every frame allows for parallelization. This scalable offline-solver can be used to generate ground truth datapoints for training real-time regressor models (see Section 9).
\subsection{Lip-Sync Deepfakes}
One well-known limitation of most motion tracking pipelines is that they struggle to capture the subtle lip/mouth movements required for a high fidelity sync to an audio track. 
To address this issue, we utilize a specialized neural network trained to inference images of the lip/mouth from speech signal input. Specifically, we use the pretrained Wav2Lip model \cite{wav2lip}, which can be used to obtain 2D image animations of lip/mouth movements given only an audio clip and a single reference image of a (realistically-textured) synthetic face in the neutral pose. One advantage of using such 2D lipsync models is that the data required to train them, i.e. videos of people talking, is much more commonly available than data for 3D talking heads. Once again, it is worth noting that we leverage an off-the-shelf pretrained neural network, and thus are able to obtain improved results as research on such networks progresses.
\par Following this approach leads to two separate videos: our segmented deepfake obtained from inferencing the source video input, and a frontal video of lip/mouth motion obtained from inferencing the source audio input. Notably, we use inverse rendering separately on each video, and subsequently blend the two sets of animation parameters together. This allows one to focus on the rigid pose and expression parameters in one video without worrying too much about the fidelity of the subtle lip/mouth motions, while focusing on highly detailed lip/mouth motion in the other video with fixed pose and (non-lip/mouth) expression  (i.e. expressionless, and in the front-facing neutral pose). Essentially, we decompose a difficult problem into two less difficult components, which can each benefit from known constraints and importance metrics. See Figure \ref{fig:lipsync}.
\figcaption{figures/figure-audio-deepfake.png}{Our proposed approach to enhance subtle lip/mouth movements. In the first row, we generate our segmented deepfake video (obtained from inferencing the source video input) and run inverse rendering to obtain control parameters. In the second row, we generate a frontal video of lip/mouth motion (obtained from inferencing the source audio input) and again run inverse rendering to obtain control parameters. Finally, we blend parameters for rigid pose and (non-lip/mouth) expressions from the first row with parameters for lip/mouth motion from the second row.}{fig:lipsync}
\section{Learned regression}
In this Section, we present work on using the learned encoder $E$ of a deepfake network in order to create a personalized regression model for motion tracking. As discussed in Section 7, we first create a synthetic dataset that well-mimics the in-the-wild data we wish to track (as measured in the deepfake latent space). Then, given ground truth synthetic parameters $p^{(i)},y^{(i)},\vec{w}^{(i)}$ paired with generated images $I_S=F_{seg}(p^{(i)},y^{(i)},\vec{w}^{(i)})$, a straightforward approach would be to train a regressor that goes from the latent embeddings of synthetic renders $E(I_S)$ to ground truth  $p^{(i)},y^{(i)},\vec{w}^{(i)}$; subsequently, that regressor could be inferenced on real-image embeddings $E(I_R)$ in order to estimate motion parameters for $I_R$. 
\par In order to reduce the domain gap even further, we project both the synthetic data and the in-the-wild images into a `synthetic' embedding space, training and inferencing on $\vec{e}=E(D_S(E(I_S)))$ and $\vec{e}=E(D_S(E(I_R)))$ instead of $E(I_S)$ and $E(I_R)$. Moreover, in order to incorporate in-the-wild images during training, we use a pretrained landmark detection network $\hat{L}$ (see \cite{bulat2017far}) to generate sparse landmarks $\vec{m}=\hat{L}(D_S(E(I_S)))$ and $\vec{m}=\hat{L}(D_S(E(I_R)))$ for weak supervision (used only during training).  With pose denoted as $\vec{p}=(p,y)$, the synthetic dataset $\Omega_S$ contains datapoints of the form $(\vec{e},\vec{p},\vec{w},\vec{m})$, while the in-the-wild dataset $\Omega_R$ contains datapoints of the form $(\vec{e},\vec{m})$ lacking ground truth pose and control parameters.
\par We designed our network model to mimic traditional solvers, with modules \textit{sequentially} predicting pose $\vec{p}$, jaw control parameters $\vec{w}_j$ (a subset of $\vec{w}$), and the remaining control parameters $\vec{w}_c$ (which we will refer to as the expression controls). Given a latent embedding $\vec{e}$, we denote the pose network as $P(\vec{e}),$ the jaw network as $J(\vec{e},P(\vec{e})),$ the expression network as $W(\vec{e},P(\vec{e}),J(\vec{e}))$, and our landmark prediction network as $L(P(\vec{e}),J(\vec{e}),W(\vec{e}))$. $P,J,W,L$ are all multi-layer perceptron networks, composed of multiple ``fully-connected / leaky ReLU" stacks with a sigmoid activation layer. Our network model is trained in three stages as illustrated in Figure \ref{fig:regressor}.

  \figcaption{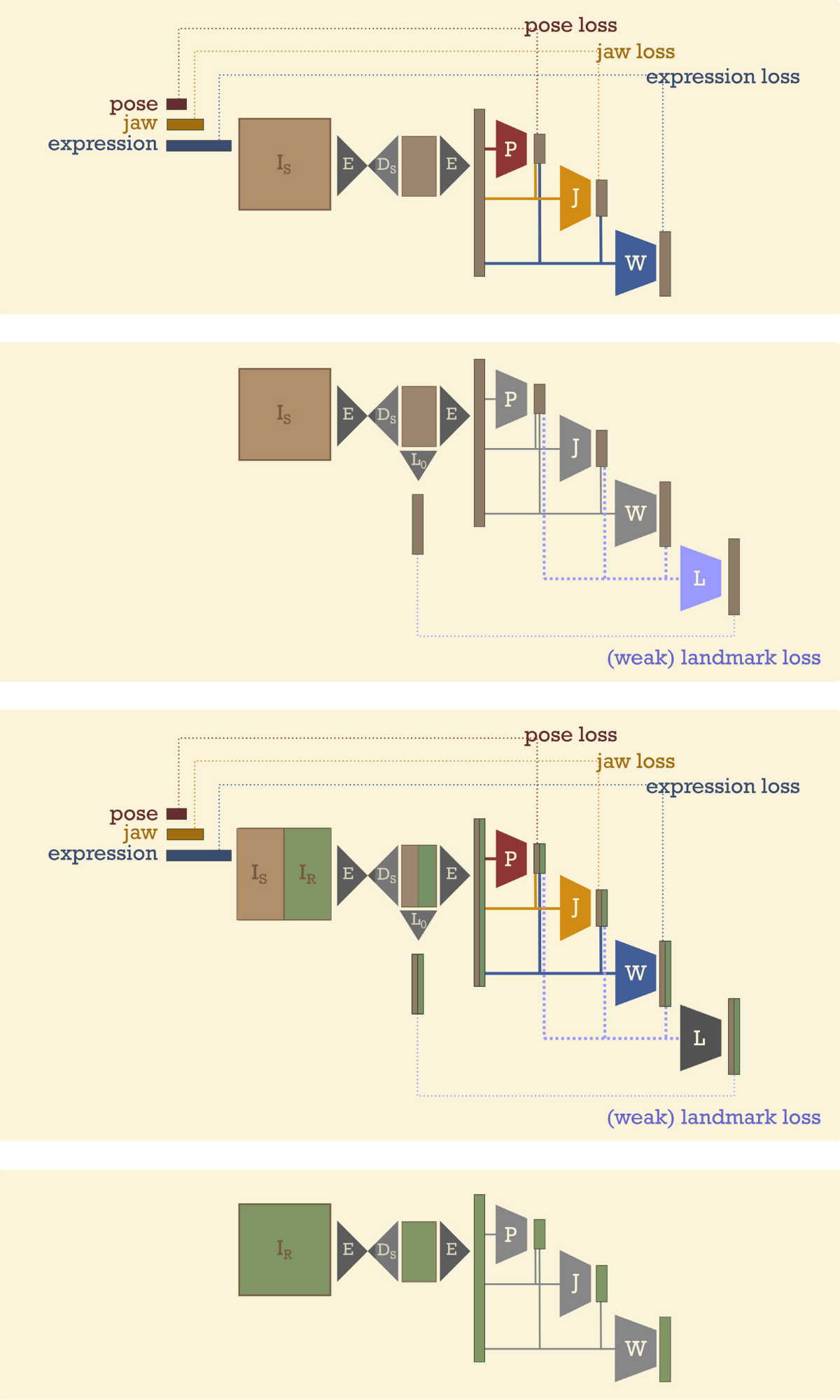}{Diagram of our regressor architecture, with the three steps of training shown top to bottom (followed by the evaluation step). Grey denotes frozen layers. The pose, jaw, and expression losses are driven only by the synthetic data (shown in brown). The in-the-wild data (shown in green) is incorporated purely through weak supervision, achieved via landmarks.}{fig:regressor}

\subsection{Training}
In the first stage, $P$, $J$, and $W$ are trained using only synthetically generated data with known ground truth (Figure \ref{fig:regressor}, top). $P(\vec{e})$ learns to predict $\vec{p}$ directly from $\vec{e},$ $J(\vec{e},P(\vec{e}))$ learns to predict $\vec{w}_j$ from $\vec{e}$ and $P(\vec{e})$, and  $W(\vec{e},P(\vec{e}),J(\vec{e}))$ learns to predict $\vec{w}_c$ from $\vec{e}$, $P(\vec{e})$, and $J(\vec{e})$. During training, the loss for $J$ not only compares $J(\vec{e},P(\vec{e}))$ to $\vec{w}_j$ as expected but also contains an equivalent term using the ground truth $\vec{p}$ in place of $P(\vec{e})$, i.e. 
$$\mathcal{L}_{J}=\sum_{\Omega_S}\nrm{J(\vec{e},P(\vec{e}))-\vec{w}_j} + \nrm{J(\vec{e},\vec{p})-\vec{w}_j}.$$
Similarly,
$$\mathcal{L}_{W}=\sum_{\Omega_S}\nrm{W(\vec{e},P(\vec{e}),J(\vec{e}))-\vec{w}_c} + \nrm{W(\vec{e},\vec{p},\vec{w}_j)-\vec{w}_c}.
$$
\par In the second stage, $P,J,W$ are frozen while we train a minimal landmark network $L$ that learns to loosely predict the estimates $\vec{m}=\hat{L}(D_S(E(I_S)))$ again using only synthetic data (Figure \ref{fig:regressor}, second row). 
To account for noisy landmarks, which is a known issue in ``ground truth" annotations \cite{dong2018supervision} and compounded by our choice of solver (see the discussion in Section 7.2), we use a loss that vanishes when the distance between each marker prediction $L_k$ and its corresponding estimate $\vec{m}_k$ is less than some threshold $\delta>0$, i.e. 
\eqnsplit{\mathcal{L}_{L}=\sum_{\Omega_S}\sum_k&\max (0,\nrm{L_k(P(\vec{e}),J(\vec{e}),W(\vec{e})) - \vec{m}_k}-\delta)\\+&\max (0,\nrm{L_k(\vec{p},\vec{w}_j,\vec{w}_c) - \vec{m}_k}-\delta).}
In our experiments, we used $\delta=0.01$ (with ground truth landmarks normalized between 0 and 1).
\par In the third stage, $P,J,W$ are fine-tuned with $L$ frozen, utilizing both the synthetic and in-the-wild data (Figure \ref{fig:regressor}, third row). 
Although $\mathcal{L}_P,\mathcal{L}_J,\mathcal{L}_W$ retain their original form, we no longer use the second term in $\mathcal{L}_L$ since it is unavailable on in-the-wild data. One could still include that second term for $\Omega_S$, but we prefer to treat $\Omega_R$ and $\Omega_S$ more similarly, and modify the landmark loss to be
\eqnsplit{\mathcal{L}^*_{L}&=\sum_{\Omega_S\hspace{1px}\cup\hspace{1px}\Omega_R}\sum_k\max (0,\nrm{L_k(P(\vec{e}),J(\vec{e}),W(\vec{e})) - \vec{m}_k}-\delta).}
 As shown in Figure  \ref{fig:ablation}
, this third training step is crucial and greatly improves the quality of the results.
 \figcaption{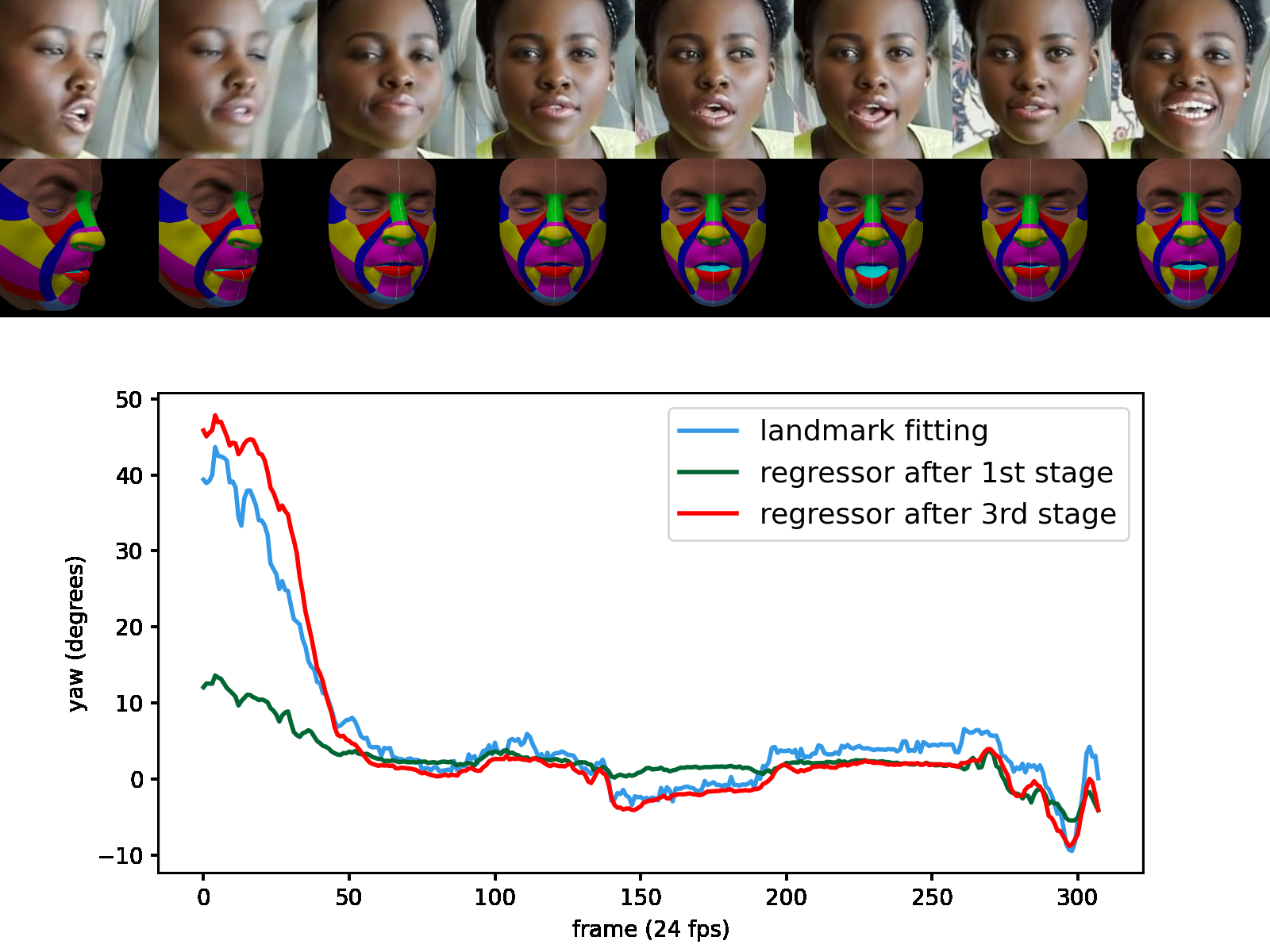}{Yaw estimation plotted across a single video of 300$+$ frames (six sample frames are shown for both the video and the corresponding output from the regressor). The predictions obtained using sparse landmarks (blue) are extremely noisy (as compared to the regressors). The 1st stage regressor (green) trained on only synthetic data struggles to accurately predict yaw on poses that are too far from the neutral, which is not the case for the final regressor (red) trained with the aid of weak supervision via landmarks.}{fig:ablation}
\subsection{Results and Discussion} Figure \ref{fig:regressorresults} shows results on three separate subjects, with no ground truth  parameters (obtained or computed).
\figcaption{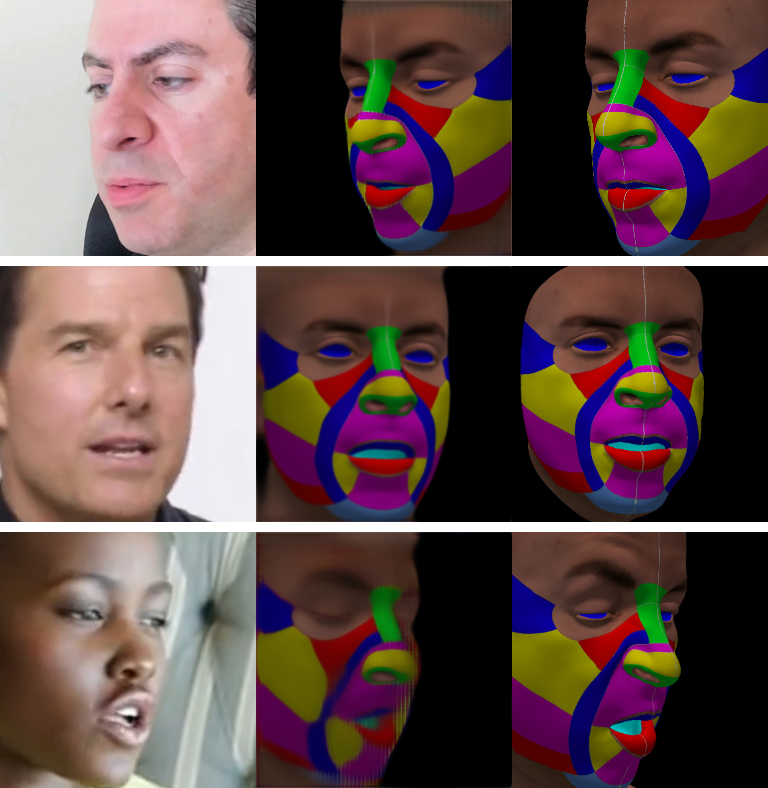}{From left to right (three examples shown): input frame $I_R$, synthetic deepfake $D_S(E(I_R))$, rendering of the 3D model with pose, jaw, and expression regressed from $E(D_S(E(I_R)))$. Interestingly, despite the degradation in image quality of $D_S(E(I_R))$, our regressor still outputs reasonable results.}{fig:regressorresults} For each example, a deepfake model was trained from scratch using a few thousand in-the-wild images as well as a similarly sized synthetic dataset (see Section 7) at $256\times256$ image resolution. The resulting deepfake model was then used to extract 512-dimensional latent embeddings from both datasets, and a regressor was trained to predict motion capture parameters from these latent embeddings. Importantly, the low-dimensionality of both the embeddings and the motion capture parameters enables quite efficient training of the regressor (only requiring minutes on a single GPU). The bottleneck is the 6-8 hours required to train a deepfake model, which is something we expect that the deepfake community will improve upon.
\par The resolution limitations of the deepfake technology similarly limit the fidelity of our results, in particular in the lip region; thus, region-specific encodings (such as the ones described in Section 8.1) are a promising avenue of future research. A deeper investigation into and a comparison of various animation rig parametrizations would also be interesting. Finally, although we propose an alternative to large models and large scale data collection, a hybrid approach that leverages pretrained models might help to alleviate the training time needed for the deepfake networks.
\section{Summary} In this paper, we presented various techniques that utilize unsupervised autoencoder deep-fake neural networks in order to create personalized facial appearance and motion capture pipelines.
Notably, only a minimal amount of subject-specific in-the-wild imagery is required, since we were able to leverage synthetically created “ground truth” data during training. Such personalized facial appearance and motion capture pipelines bypass issues with bias that plague large-scale data collection and large pre-trained monolithic models. Our new approach has obvious potential benefits beyond appearance/motion capture and retargeting. For example, the reliance on only subject-specific in-the-wild data could be leveraged to parametrize and analyze the facial motion present in videos in order to detect frame based video forgery (e.g. as created by deepfakes).

\section{Acknowledgments}
This work was supported by a grant from Meta (née Facebook) and in part by ONR N00014-13-1-0346, ONR N00014-17-1-2174. We would also like to thank Industrial Light and Magic for supporting our initial experiments on deepfake based facial capture, Robert Huang and William Tsu for their kind donation of an NVIDIA Titan X GPU which was used to run experiments, and Epic Games for their help with the Metahuman rig.

{\small
\bibliographystyle{packages/ieee}
\bibliography{references}
}
\end{document}